%% file: aaai25.tex
\documentclass[letterpaper]{article} 
\usepackage{aaai25}  
\usepackage{times}  
\usepackage{helvet}  
\usepackage{courier}  
\usepackage[hyphens]{url}  
\usepackage{graphicx} 
\usepackage{natbib}  
\usepackage{caption} 
\usepackage{booktabs}
\usepackage{xspace}
\usepackage{multirow}

\usepackage{bm}
\usepackage{algorithm}
\usepackage{algpseudocode}
\usepackage{algorithmicx}
 \usepackage{amsmath} 
 \usepackage{amssymb} 
\usepackage{gensymb}
\usepackage{subcaption}
\usepackage{xcolor}
\usepackage{hyperref}

\usepackage{newfloat}
\usepackage{listings}
\DeclareCaptionStyle{ruled}{labelfont=normalfont,labelsep=colon,strut=off} 
\floatstyle{ruled}
\newfloat{listing}{tb}{lst}{}
\floatname{listing}{Listing}
\title{CAD-NeRF: Learning NeRFs from Uncalibrated Few-view Images \\by CAD Model Retrieval}
\author{
    Xin Wen, Xuening Zhu, Renjiao Yi\equalcontrib, Zhifeng Wang, Chenyang Zhu, Kai Xu\equalcontrib
}
\affiliations{
    National University of Defense Technology\\
    Changsha, 410073, China
}

\usepackage{bibentry}

\begin{document}

\input{hyphen}
\maketitle

\begin{abstract}
Reconstructing from multi-view images is a longstanding problem in 3D vision, where neural radiance fields (NeRFs) have shown great potential and get realistic rendered images of novel views. Currently, most NeRF methods either require accurate camera poses or a large number of input images, or even both. Reconstructing NeRF from few-view images without poses is challenging and highly ill-posed. To address this problem, we propose CAD-NeRF, a method reconstructed from less than 10 images without any known poses. Specifically, we build a mini library of several CAD models from ShapeNet and render them from many random views. Given sparse-view input images, we run a model and pose retrieval from the library, to get a model with similar shapes, serving as the density supervision and pose initializations. Here we propose a multi-view pose retrieval method to avoid pose conflicts among views, which is a new and unseen problem in uncalibrated NeRF methods. Then, the geometry of the object is trained by the CAD guidance. The deformation of the density field and camera poses are optimized jointly. Then texture and density are trained and fine-tuned as well. All training phases are in self-supervised manners. 
Comprehensive evaluations of synthetic and real images show that CAD-NeRF successfully learns accurate densities with a large deformation from retrieved CAD models, showing the generalization abilities.  
\end{abstract}

\input{intro}

\input{related}

\input{method}

\input{results}


\input{conclusion}
\clearpage
\bibliography{aaai25}

\clearpage
\input{supp}

\end{document}

%% file: hyphen.tex
\hyphenation{sinusoi-dal ac-cepts acc-epts Trans-former Co-FiNet CoFi-Net sce-narios scena-rios domi-nated Al-though Alth-ough gene-rate genera-te hie-rarchical em-bedding Geo-metric gene-rates}

\def\todo#1{\textcolor{red}{#1}}
\def\secname#1{\textcolor{red}{#1}}
\def\head#1{\noindent\textbf{#1}}
\def\tight#1{\hspace{1pt}{#1}{\hspace{1pt}}}
\def\medium#1{\hspace{2pt}{#1}{\hspace{2pt}}}
\def\wide#1{\hspace{3pt}{#1}{\hspace{3pt}}}
\newcommand{\slfrac}[2]{\left.#1\middle/#2\right.}
\newcommand{\tabincell}[2]{\begin{tabular}{@{}#1@{}}#2\end{tabular}}
\newlength{\wdth}
\newcommand{\strike}[1]{\settowidth{\wdth}{#1}\rlap{\rule[.5ex]{\wdth}{.4pt}}#1}
\newcommand{\ptitle}[1]{\noindent\textbf{#1}\hspace{5pt}}

\newcommand{\kx}[1]{{\color{red}#1}}
\newcommand{\tofix}[1]{{\color{red}[Fix: #1]}}
\newcommand{\num}{{\color{blue}[X]~}}
\newcommand{\supl}[1]{{\color{black}\emph{#1}}}
\newcommand{\fix}[1]{{\color{blue}#1}}

%% file: intro.tex
\section{Introduction}
\begin{figure}[t]
\centering
  \includegraphics[width=1.0\linewidth]{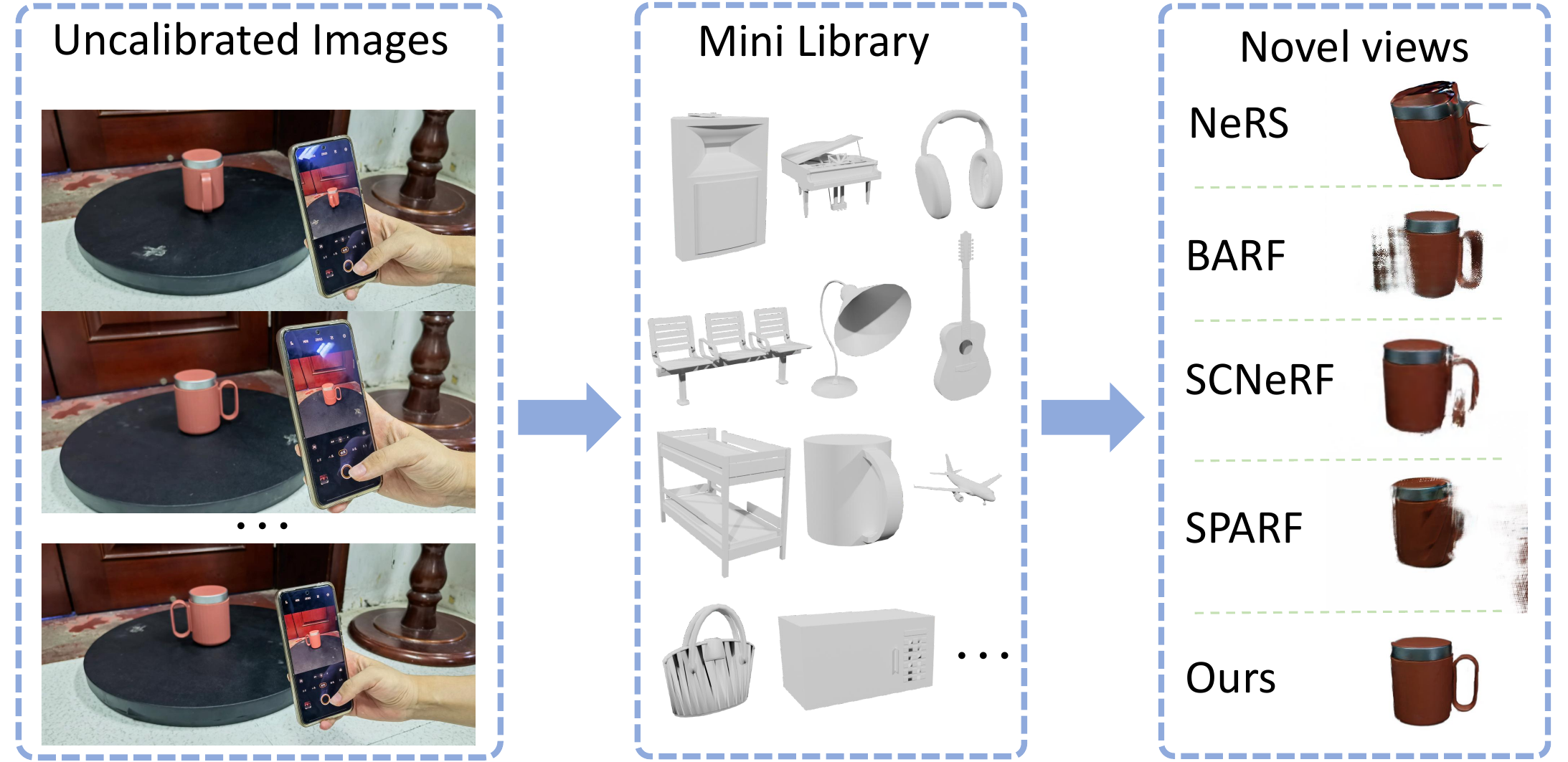}
  \caption{CAD-NeRF takes few-view images with unknown poses as inputs, and jointly optimizes density, texture, and poses with the help of priors from the CAD mini library. }
  \label{fig:teaser}
  \vspace{-0.2cm}
\end{figure}

Reconstruction and synthesizing new views of 3D scenes from 2D images is a core problem in the field of 3D vision. With the development of differentiable rendering, NeRF~\cite{2021nerf} and its subsequent works \cite{tancik2022block,barron2021mip,nerfinwild,sun2024recent,yan2024dialoguenerf} have shown great potential in this task, with implicit representations of multi-layer perceptrons (MLPs) to represent 3D objects or scenes. 
NeRFs succeed in generating high-quality and realistic images of novel views but with some limitations. 
The limitations include the requirement of a large number of multi-view images (around 100 images for a scene), and the assumption of accurate camera poses. Meanwhile, each scene is trained from scratch which takes a long time to converge. They highly limit the practicality of NeRFs. 

Recent works~\cite{chen2021mvsnerf,yu2021pixelnerf,tancik2021metanerf,jain2021putting,zhang2021ners,wang2021ibrnet,deng2022depth} explore to address the problem of novel view synthesis from few views. Most of them learn camera poses by pretrained networks. They are trained from several multi-view image datasets, which is time-consuming. DietNeRF~\cite{jain2021putting} takes advantage of semantic information using a pretrained visual encoder CLIP, to improve the few-shot quality. However, it does not pay attention to the geometry of the object, which may fail when the geometry is complicated or very different from the training set. DS-NeRF~\cite{deng2022depth} proposes a depth loss to learn better geometry that utilizes depth supervision. Most importantly, all of these few-image works require accurate camera poses of input images. To solve this problem, NeRS~\cite{zhang2021ners} softens the constraint of input poses by adding pose refinement in the pipeline, where they only require rough camera poses from fixed views as inputs. NeRS demonstrates that the method can be better applied to real scenes and used by amateur users without the requirement of accurate poses.  

In this paper, we aim to explore the reconstruction from a small number ($<10$) of uncalibrated images. As illustrated in Figure~\ref{fig:teaser}, we propose CAD-NeRF, a NeRF-based method that learns geometry and texture representations of objects from 3-10 uncalibrated images. The problem is challenging since camera poses are completely unknown, and input views are extremely few. To tackle the problem, we first construct a mini library of common objects from ShapeNet~\cite{chang2015shapenet}. Each object is pre-rendered from uniformly sampled poses. By the library, we can retrieve an initial pose and similar mesh for each image by comparing the silhouettes. We first retrieve a most similar mesh by voting of multiple input images and then we retrieve poses for input images. In this process, we find searching for a pose with the most similar silhouette for each input image may lead to problems of pose conflicts among views. 
Due to shape symmetries, some retrieved poses may have errors of 180 degrees. The retrieved poses will not follow the initial order of input images. To maintain the image order and get a coherent pose order, we propose a multi-view pose retrieval method that considers the order information of input images by utilizing the backtracking algorithm to search multi-view poses at the same time. The retrieved mesh serves as the supervision when learning the density fields. Later, the deformation of the density field and camera poses are trained jointly by a re-rendering loss of the rendered images. At last, the texture and density are trained and fine-tuned by the re-rendering loss of the rendered color images.  
Our main contributions are:
\begin{itemize}
\item We propose a method that reconstructs NeRFs from very few images without camera poses, with the help of prior shapes from CAD datasets. From the results, objects that are very different from CAD models can still be successfully reconstructed. The method effectively boosts the generalization ability of NeRFs. 
\item We propose a multi-view pose retrieval method, that considers the ordering of the input images, to prevent pose conflicts of input images. It is a new problem in NeRF-based methods.  
\item On both synthetic and real objects, CAD-NeRF demonstrates significant improvement over state-of-the-art and NeRF baselines. Since the proposed method is not keypoint-based, it works for feature-sparse cases which is challenging for other methods. 
\end{itemize}

%% file: related.tex
\begin{figure*}[t]
\centering
  \includegraphics[width=1.0\linewidth]{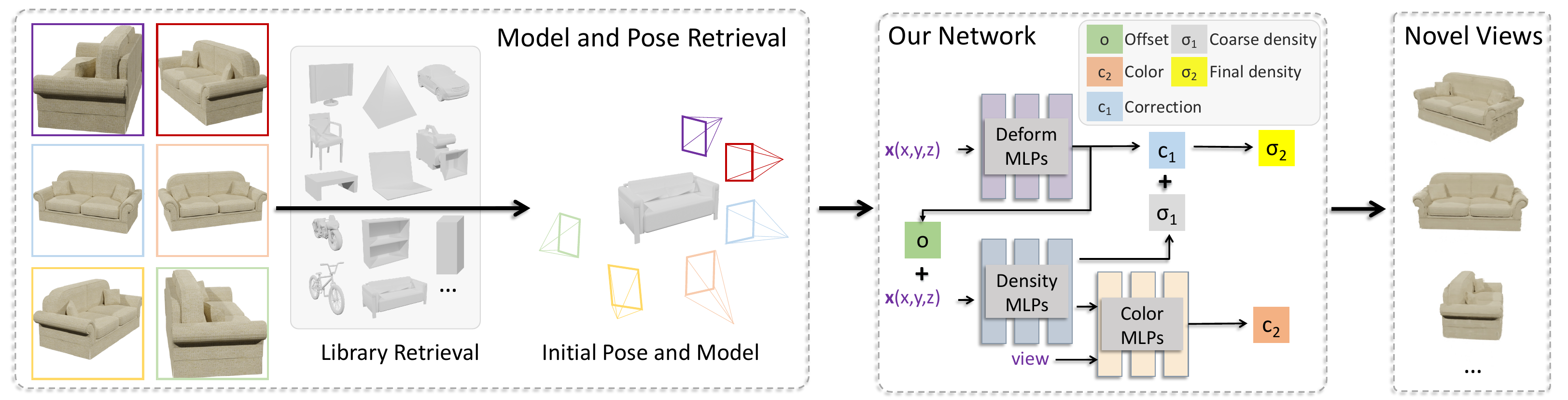}
  \caption{The CAD-NeRF pipeline. Input images are used to retrieve the model of poses from the mini library. The CAD model is treated as the supervision to pre-train the initial density field (phase one). In phase two, sampled rays are sent from retrieved poses to generate 3D points along the rays, for the deformation network to predict the offset and correction of each point. Poses are optimized at the same time. In phase three, a color network is added and three networks are trained together. 
  }
  \label{fig:overview}
\end{figure*}

\section{Related works}
\label{sec:related}

\ptitle{Neural representations.} There are various 3D representations used in 3D reconstruction approaches. Previous methods mainly use supervised training on explicit representations including voxel grids~\cite{choy20163d,girdhar2016learning,yu2021framework}, meshes~\cite{gkioxari2019mesh,wang2018pixel2mesh,feng2018perceptual}, point clouds~\cite{yang2019pointflow,fan2017point,tian2023tensorformer,tian2024surface,sun2024review}. NeRFs~\cite{2021nerf} achieved state-of-the-art performance on this task by implicit representation and volume rendering. 
Many follow-up approaches based on NeRFs make improvements in different aspects, such as faster inference~\cite{reiser2021kilonerf,lin2022enerf}, unconstrained images \cite{nerfinwild}, generalization ~\cite{yu2021pixelnerf,xu2022point}, lighting ~\cite{zhu2024relighting} and pose estimation~\cite{cheng2023lu,lin2021barf,hamdi2022sparf,fan2024revisit}. BARF~\cite{lin2021barf} learning 3D scene representations by jointly optimizing the radiance field and camera parameters, but they need initial noisy camera poses. SCNeRF \cite{jeong2021self} proposes a novel geometric loss to learn more accurate camera parameters. SPARF~\cite{hamdi2022sparf} shows great performance on sparse input images with noisy camera poses. They propose a correspondence network to obtain extra information from the input views and the novel depth loss to guarantee view consistency. Different from these methods, we train from uncalibrated images and initialize poses by retrieving them from the CAD library. 

\ptitle{Novel view synthesis.} Novel view synthesis methods usually require a large number of input images. Recent works~\cite{chen2021mvsnerf,yu2021pixelnerf,tancik2021metanerf,jain2021putting,zhang2021ners,wang2021ibrnet,deng2022depth,wu2019joint,niemeyer2022regnerf,kim2022infonerf,yuan2022neural,liu2022efficient, gao2024generic} have made rapid progress towards few-image inputs. Most of them still need to be pretrained on multi-view image datasets of various scenes, which is time-consuming, or requires extra depth information. DS-NeRF~\cite{deng2022depth} proposes a depth loss for learning better geometry that utilizes depth supervision, but it works better on data with rich textures. For objects with simple textures, it is difficult to match feature points due to sparse features, thus hindering the generation of sparse 3D points. In addition, these few-image works require the input images to have accurate camera poses, which may be infeasible to get in the real world. Differing from these works, NeRS ~\cite{zhang2021ners} provides the initial mesh to learn the neural shape representation of the surface while ensuring a watertight reconstruction. However, it is limited to real objects similar to the initial mesh, and the input images have to be captured from fixed views to get initial poses. 
Unlike NeRS, our method requires very few images (3-9) that are fully uncalibrated and captured from random views. It enables amateur users to take photos on mobile devices. 

\ptitle{Shape deformation and 3D CAD model retrieval.} Shape deformation on implicit representation is a new problem. Deng~\cite{deng2021deformed} proposes a deformation implicit field (DIF) representation 
for modeling 3D shape instances of each category, 
which is mainly used for texture transfer and shape editing. Inspired by DIF, we design a method based on NeRFs to utilize the deformation of the initial mesh, to achieve the purpose of effectively constructing NeRFs with few-view images. Different from their purpose of surface reconstruction, we adopt classic volume rendering for object reconstruction. Moreover, DIF is leveraged for the deformation of one category of objects, where we build a CAD library of common objects and retrieve the most similar ones, without the requirement of object categories. In addition, CAD model retrieval is also an important research problem with many studies~ \cite{qin2018fine,qin20223d,hou2023fus}, Qin et al. \cite{qin20223d} utilizes sketches, unsupervised learning method, and recursive neural network-based deep variational autoencoders to effectively match models in a feature space, validated through experiments on a dataset of approximately 2,000 models. A new approach for CAD model classification and retrieval \cite{hou2023fus}, utilizing a B-rep graph descriptor and the FuS-GCN neural network with a fusion self-attention graph pooling algorithm, is proposed to effectively extract topological and geometric features, outperforming existing 3D shape descriptors like point clouds, voxels, and meshes. Unlike these methods, the proposed method is not based on neural networks but employs a simple and efficient 2D image-matching method to search for CAD models. Since we do not require very precise matches, similar models are sufficient to help us optimize pose and initialize the network. Furthermore, we propose the multi-view pose retrieval method to get more accurate poses.

%% file: method.tex
\section{Method}

In this section, we introduce the detailed technical method of CAD-NeRF, whose pipeline is illustrated in Figure~\ref{fig:overview}. First, a brief review of volumetric rendering utilized by NeRF representation is introduced in Section 3.1. The multi-view model and pose retrievals are introduced in Section 3.2.
For joint training of density, texture, and poses, we introduce the pretraining of the density field in Section 3.3, as phase one. In phase two, the deformation of the density field and camera poses are optimized jointly at the same time, as introduced in Section 3.3 and 3.4. In phase three, a color network is added to learn the textures, and the whole pipeline is trained simultaneously, as introduced in Section 3.5. Additionally, the pipeline of training images with backgrounds is introduced in Section 3.6.

\subsection{Volumetric rendering}
\label{sec:nerf}

Given a large set of multi-view images with accurate camera poses, NeRF~\cite{2021nerf} learns to represent the scene as a continuous radiance field of color and density, to render novel views by the differentiable volumetric rendering layer. 
Specifically, given a 3D point $\mathbf{x} \in \mathbb{R}^{3}$ and a viewing direction $\mathbf{d} \in \mathbb {R}^2$, NeRF learns an implicit function $f$ to represent the density $\sigma$ and RGB color $\mathbf{c}$ as $f(\mathbf{x}, \mathbf{d}) = (\sigma, \mathbf{c})$. Given a camera pose $\mathbf{P}$, the camera ray is defined as: $\mathbf{r}(t) = \mathbf{o}+t\mathbf{d}$, where $\mathbf{o}$ is the camera origin. Here the near and far bounds are defined as $t_{n}$ and $t_{f}$, and the rendered image from pose $\mathbf{P}$ can be computed by:
\begin{equation}
 \hat{\mathbf{C}}(r)=\int_{t_{n}}^{t_{f}} T(t) \sigma(t) {\mathbf{c}}(t) d t,
\end{equation}
where $T(t)=\exp \left(-\int_{t_{n}}^{t} \sigma({r}(s)) d s\right)$. 
To train NeRF without 3D supervision, we can define a re-rendering loss between the real and re-rendered images from pose $\mathbf{P}$. The re-rendering loss is defined as:
\begin{equation}
\mathcal{L}_{color}= \sum_{\mathbf{r} \in \mathcal{R(\mathbf{P})}}\|\mathbf{C}(\mathbf{r})-\hat{\mathbf{C}}(\mathbf{r})\|_2^2, 
\end{equation}
where $\hat{\mathbf{C}}$ and ${\mathbf{C}}$ denote the rendered color and ground truth color respectively. $\mathcal{R}(\mathbf{P})$ is the set of rays of target pose $\mathbf{P}$.

NeRF achieves extraordinary performance on the task of novel view synthesis, and it becomes a milestone in 3D vision. It draws attention to the implicit representation of 3D scenes. However, the original NeRF has several main limitations. It needs a large number (usually around 100) of images as inputs, and the training is time-consuming and computationally expensive. Each input image needs accurate poses, which prevents us from utilizing numerous uncalibrated images in real life and on the Internet. Our method builds upon NeRF but tries to soften these limitations. 

\subsection{Library retrieval}
\label{sec:retri}

\ptitle{CAD model retrieval. }

The task of reconstructing NeRF from few-view images with unknown poses is challenging for two reasons. 
On one hand, for a single object, it is difficult to use COLMAP ~\cite{schoenberger2016sfm} to obtain camera poses from such a small number of ($<10$) images, especially when the object is smooth in textures and sparse in features. Features are too few to match. In most cases, COLMAP would fail to estimate reasonable poses for our data. Thus, estimating camera poses via COLMAP is infeasible here. 
On the other hand, camera poses are hard to obtain for amateur users. Without accurate poses, NeRF suffers from bad initializations and fails to work.  

Therefore, with a large amount of CAD models out there, we start to consider how to make use of them to solve the problem above. 
It is hard to find an accurate CAD model for input images, but there are numerous models of a large variety. It is possible to retrieve a similar shape to serve as geometry priors. So we propose to build a mini library of CAD models and their posed renderings for retrieving initial poses and geometry.

\begin{figure}
\centering
  \includegraphics[width=1.0\linewidth]{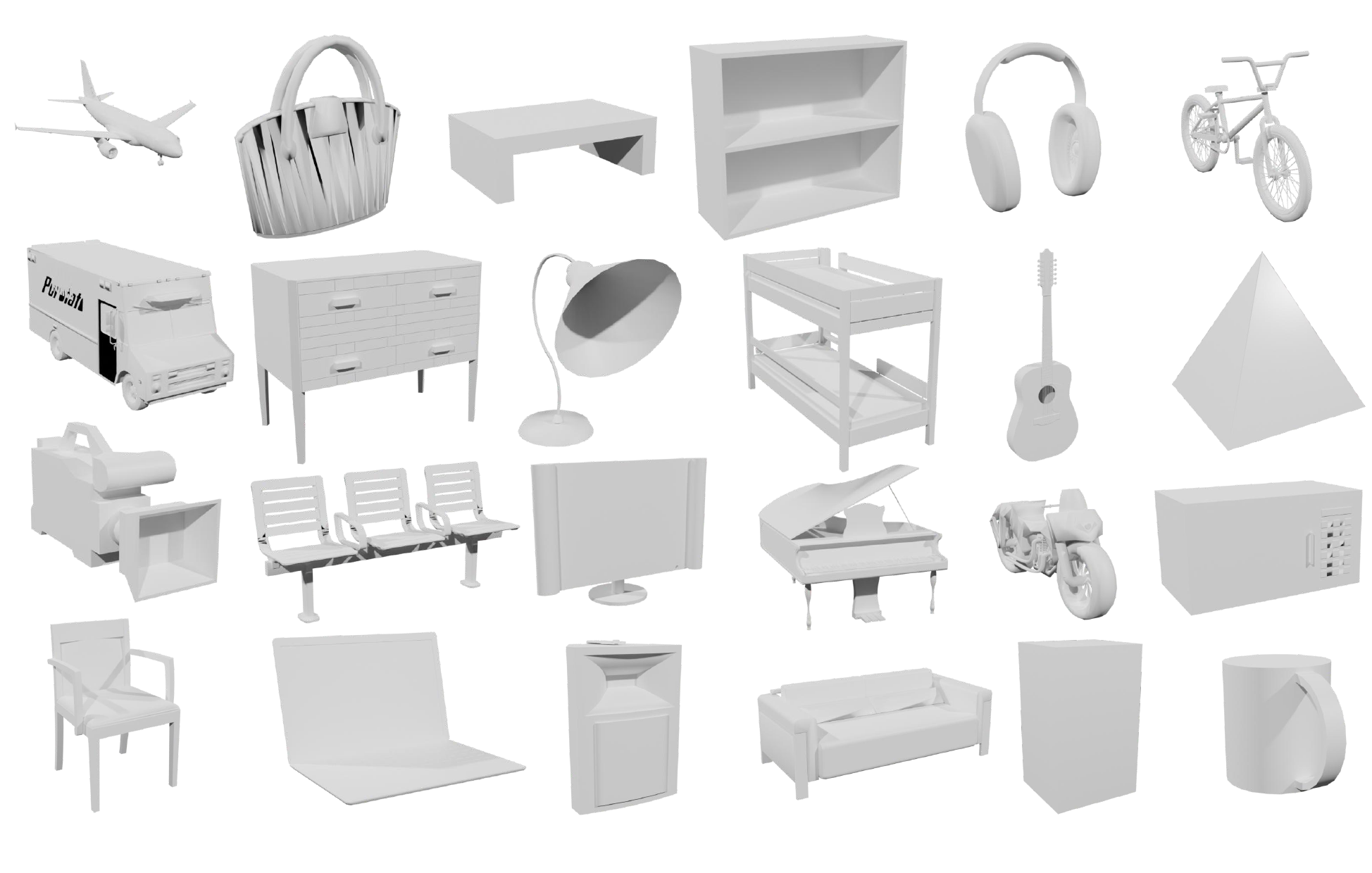} 
  \caption{CAD models in the library.}
  \label{ku}
\end{figure}

We make use of a large-scale CAD dataset ShapeNet~\cite{chang2015shapenet}, where CAD models with category labels are provided. We build a CAD library of 24 common objects, such as benches, cups, pianos, airplanes, and microwaves, trying to cover a large variety. The models are shown in Figure~\ref{ku}, 
We also add basic shapes such as cuboids, and boxes. For objects that cannot find a similar shape in those common objects, they can initialize from these basic shapes. 
We do not build a very large library because there is no need to match the exact shape of each object. The speed of retrieval is also much faster with a small library. Objects with similar shapes are good enough as shape initializations. As shown in Section 4.2, the lamp is initialized from a monitor, which is roughly matched in shape, and it is well reconstructed. 
For models in the library, We pre-render images with 100 uniformly sampled poses, which are at fixed distances and different angles. These images and the corresponding CAD models form the mini library. 

\begin{figure}
\centering
  \includegraphics[width=1.0\linewidth]{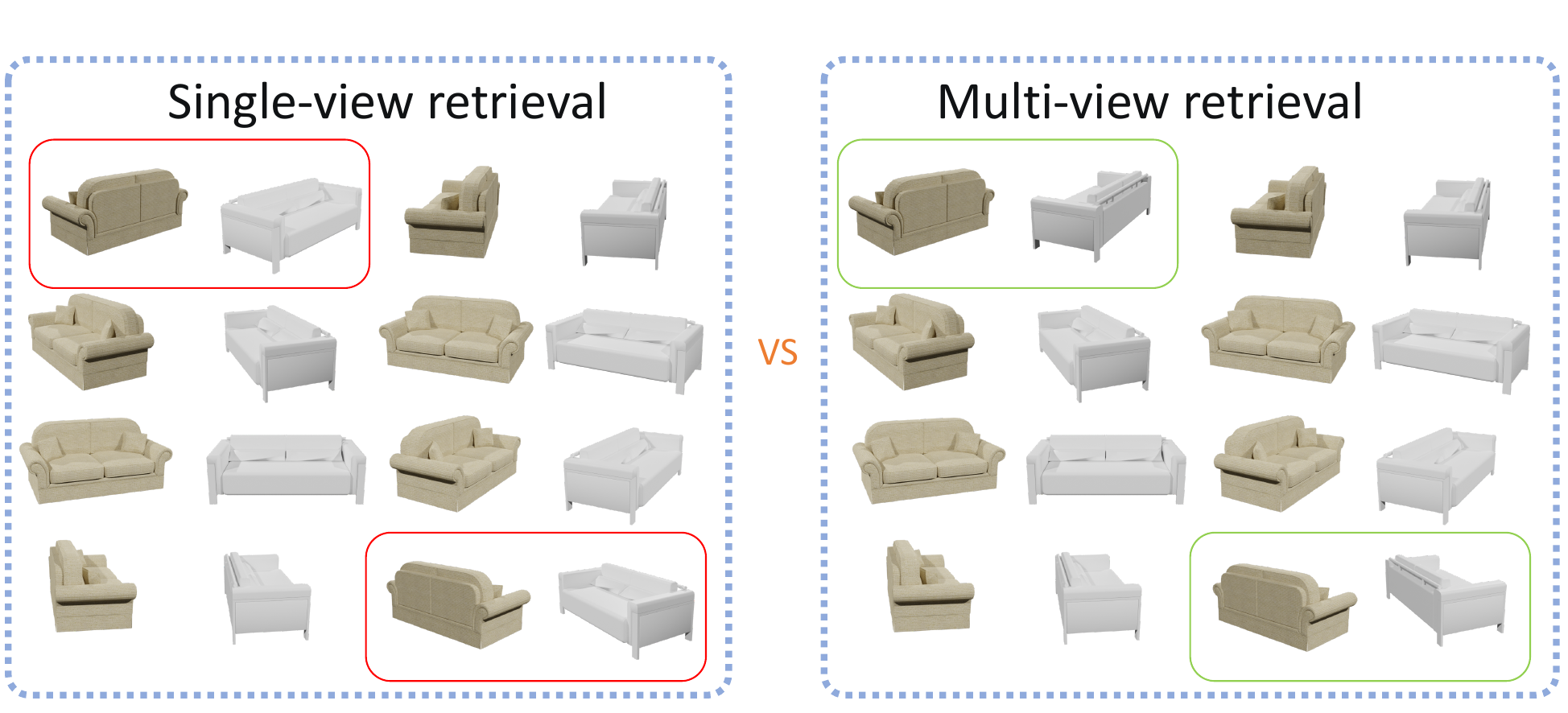}
  \caption{Comparisons of single-view retrieval and the proposed multi-view retrieval.}
  \label{compare1}
\end{figure}

\ptitle{Multi-view pose retrieval. } For model and pose retrieval, we utilize silhouette information to retrieve the closest matched image in the mini library for each input image. By calculating and comparing the score of IoUs (Intersection over Union) between mask images of pre-rendered object images in the library and input images, we find a match for each input image. To deal with the size variations, we normalize mask images by cropping the backgrounds and normalizing the length or width of objects (normalizing the larger ones in length and width). Moreover, if input images get matches from different CAD models, the one with the highest votes is selected as the retrieved model. 

In this way, the matched images are selected with the highest IoU value of this specific CAD model for each single-view image. 
Then we find a problem while applying such a strategy for multi-view images. 
Since some shapes are symmetric, the silhouettes of one view and the opposite view (add 180 degrees) may be very similar. It may lead to mismatches. For multi-view images, two close input views may be matched to opposite poses. This violates the order of input images. This problem is demonstrated in Figure~\ref{compare1} (left). 
Therefore, we consider the order information of input images and make use of their mutual relations. Since input images are very few, it is easy to rank them in order. We apply the model and pose retrieval for these images at the same time, constraining the pose order to be the same as the image order, with the maximum sum of IoUs. 
The backtracking algorithm is adopted to solve the problem. For each view/image, we no longer match the one with the highest IoU. Instead, we obtain the top $10$ corresponding matched images of each image according to IoU ranking as a candidate list $data$. For all views, we aim to find a set of correspondence respecting to the image order from the candidate lists. 
To reduce the time complexity, pruning is carried out according to whether the order conditions are met in the recursive process, for narrowing down the search scope and improving search efficiency. See Algorithm 1 for detailed steps. 
However, following the algorithm, it is still possible that it cannot get a reasonable solution satisfying the image order. In this case, 1-2 images may be discarded to get a solution for the rest images. For example, in this paper, the lamp data uses only 7 out of 9 images. 


\begin{algorithm}[tb]
\caption{Multi-view pose retrieval}
\label{alg:algorithm}
\textbf{Input}: The input $data$ is a dictionary where each key corresponds to an image, and the corresponding value consists of two lists: ``png'' and ``IoU''. The ``png'' list represents a list of input images, while the ``IoU'' list represents the IoU scores for the images corresponding to the keys in the ``png'' list.
\textbf{Parameter}: selected ordered images: sLis, current index: ind, current score: cs, current key: ck, current data: cd, max IoU value of ordered images: mLis, max score: ms.\\
\textbf{Parameter-Initialization}: sLis: [], ind: 0, cs:0, mLis: [] , ms: 0. \\
\textbf{Output}: max sequence: mLis, max score: ms.
\begin{algorithmic}[1] 
    \Function{\textbf{backtrack}}{$data, Parameter$}
        \State If all keys have been considered, return the current results: 
        \If {$ind=len(data)$}
            \If {$cs > ms$}
                \State Update the best order and score:
                \State $mLis$=copy($sLis)$, $ms=cs$.
            \EndIf
            \State \textbf{return} $mLis$ and $ms$.
        \EndIf
    \State Get the current key and its corresponding data:
    \State $ck$=$data$.keys( )[$ind$], $cd$=$data$[$ck$]. 
    \State Explore matched image from the current image:
        \For{$i$ in \{0,..., length of $cd$[png]\}}
            \State $num=cd$[png][i], $score=cd$[IoU][i].
            \If {$sLis$ is null or $num > sLis[-1][0]$}
                \State add ($num$, $ck$) to $sLis$.
                \State $mLis, ms$=BACKTRACK ($data, sLis, $
                \State $ind + 1, cs + score, mLis, ms$).
                \State remove the last element from $sLis$.
            \EndIf
        \EndFor
\State {$mLis, ms$ = BACKTRACK ($data, sLis, ind+1,$ 
\State $cs, mLis, ms$).}
\State Return the best order and score:
\State \textbf{return}  $mLis$, $ms$.
     \EndFunction
\end{algorithmic}
\end{algorithm}

It is efficient to get the closest CAD model and initial poses, without the risk of overfitting as learning-based methods. Here although our poses sampling in the mini-library is sparse, with the proposed multi-view pose retrieval, we find that average rotation errors in retrieved poses are within 15 degrees from the experiments. 

\subsection{Pretraining and deforming the density field}
\label{sec:density}
We learn an initial NeRF density field, by the retrieved CAD model. It serves as the ground truth to supervise the training of the density network. The loss function is:
\begin{equation}
\label{ldensity}
\mathcal{L}_{\text{density}}=-\frac{1}{N} \sum[\sigma \ln \sigma_t+(1-\sigma) \ln (1-\sigma_t)], 
\end{equation}
where $N$ is the number of points, $\sigma$ is the predicted density and $\sigma_t$ is the ground truth occupancy from CAD models.

To get occupancy from CAD models, We first convert CAD model files into the watertight mesh representations following ~\cite{riegler2017octnetfusion}. Then, the occupancy $\text{O}$ of any 3D point can be calculated, where $1$ indicates that the 3D point belongs to the object, and $0$ indicates the opposite:
\begin{equation}
\mathrm{O}(\mathbf{x} ): \mathbb{R} ^{3} \to [0,1].
\end{equation}
In this step, we can judge by the number of intersection points between a ray starting from point $\mathbf{x} $ and the mesh. If the number of intersection points is even, it means point $\mathbf{x}$ is located outside the mesh; otherwise, it is located inside the mesh. We set the occupancy of the points inside the mesh to be 1 and the occupancy of the points outside the mesh to be 0. The outputs of the density network are also normalized to values in $[0,1]$. The pre-training of the density field is phase one in our joint training pipeline. 

Since the retrieved model is a similar shape, not the accurate one. We further refine the density field by a deformation network $G$ from ~\cite{deng2021deformed}. It consists of two parts, which learn the offset value $\mathbf{o}\in \mathbb{R}^3$ and correction value $c_{1}\in \mathbb{R}$ of 3D point $\mathbf{x} $ respectively:
\begin{equation}
G: \mathbf{x}  \in \mathbb{R}^3 \rightarrow(\mathbf{o}, c_{1}) \in \mathbb{R}^4.
\end{equation}
By adding the offset $\mathbf{o}$ to the input point $\mathbf{x}$ coordinates $(x,y,z)$, we get augmented input coordinates. By feeding them to the density network $D$, we can get the rough density $\sigma_{1}$. Combining the rough densities with the correction values, the accurate density $\sigma_{2}$ after deformation is obtained. The deformation step can be formulated as:
\begin{equation}
\sigma_{2}=D(\mathbf{x} +\mathbf{o}) + c_{1}. 
\end{equation}


\subsection{Optimizing camera poses}\label{sec:pose}

Since the retrieved mesh and poses are rough ones, we continue to optimize camera poses while deforming the density fields. It performs as phase two, 
and we adopt coarse-to-fine manner and positional encoding as NeRF \cite{2021nerf} does. 
We parameterize the rotation angle of the camera $\mathbf{R}$ and the translation distance $\mathbf{t}$ following to \cite{wang2021nerfmm}, and convert them into a transformation matrix $\mathbf{T}$. By multiplying the transformation matrix $\mathbf{T}$ with the initial pose $\mathbf{P}_{\text{init}}$, we get the optimized pose $\mathbf{P}_{\text{opt}}$:
\begin{equation}
\mathbf{P}_{\text{opt}} = \mathbf{T} \cdot \mathbf{P}_{\text{init}}
\end{equation}
The deformation network and poses are optimized simultaneously. Here, we found that due to the few number of images, continuously optimizing the pose can easily lead to overfitting. Therefore, we only optimize the pose for a certain number of iterations (e.g., from 7.5k to 10k iterations), and then fix the pose while continuing to optimize the other parts of the network. 
This phase is trained in a self-supervised manner by a re-rendering loss. 

\begin{figure}[t]
\centering
  \includegraphics[width=1.0\linewidth]{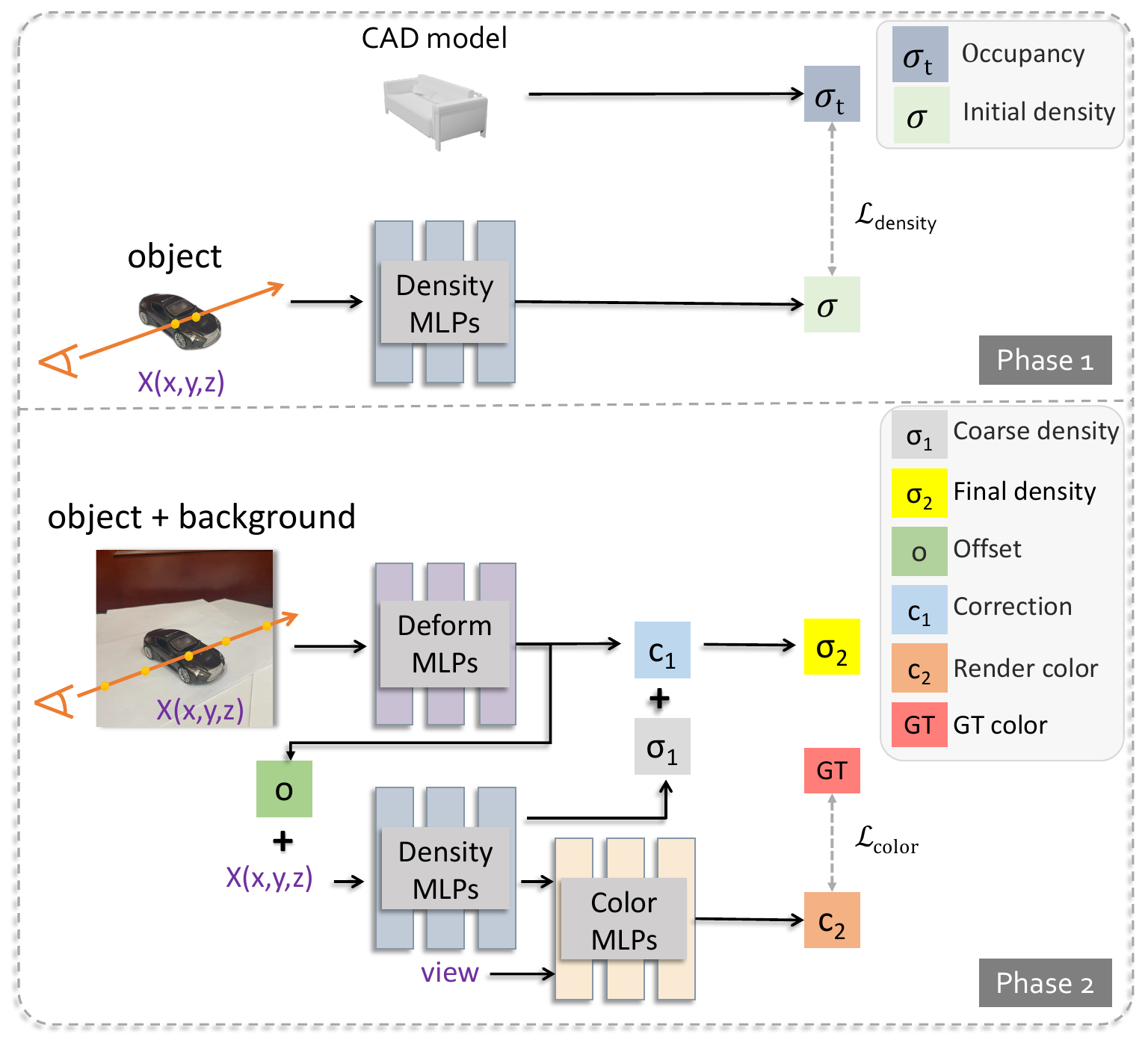}
  \caption{Training pipeline of images with backgrounds.}
  \label{tu}
\end{figure}


\subsection{Joint training of the whole pipeline}
\label{sec:joint}
After the above two phases, the density network feeds the feature layer to a color network, so that the color network can learn the relevant density information. 
Observed color may be view-dependent, so the view direction is added to the color network. Here we adopt re-rendering loss between color renderings and input images. We train the color network, fine-tune deformation, and density networks altogether simultaneously as phase three. We regularize the offset and correction to ensure that the deformation is as small as possible:

\begin{equation}
\mathcal{L}_{\text{offset}} = \sum\left\|\mathbf{o}\right\|_{2},
\mathcal{L}_{\text{correction}} = \sum\left | c_{1} \right |.
\end{equation}

In summary, the training loss of phase three is:
\begin{equation}
\label{lall}
\mathcal{L}_{\text {total }}=\mathcal{L}_{color}+\lambda_{a} \mathcal{L}_{\text{offset}}+\lambda_{b} \mathcal{L}_{\text{correction}},
\end{equation}
where $\lambda_a$, $\lambda_b$ are hyper-parameters.

\begin{figure*}[h]
\centering
  \includegraphics[width=1.0\linewidth]{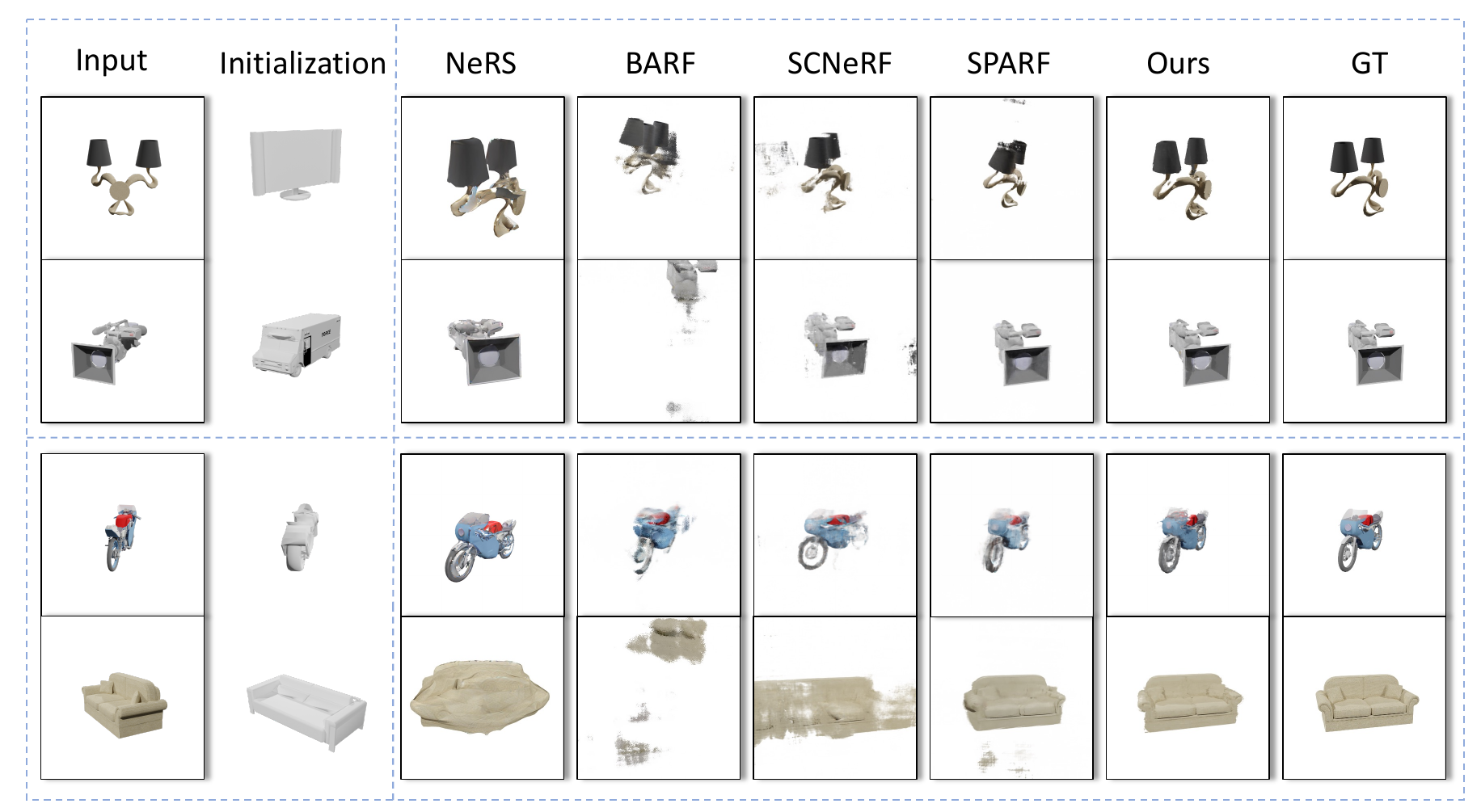}
  \caption{Qualitative comparison on ShapeNet dataset. For each data, the first column on the left is an example of the input images, here the second column is the CAD model we retrieved. On the right part, it shows novel view synthesis results of NeRS, BARF, SCNeRF, SPARF, and ours, comparing with ground truths. The numbers of uncalibrated input images for four data are 3/6/9/3, respectively.}
  \label{comp1}
\end{figure*}

\begin{figure*}[h]
\centering
  \includegraphics[width=1.0\linewidth]{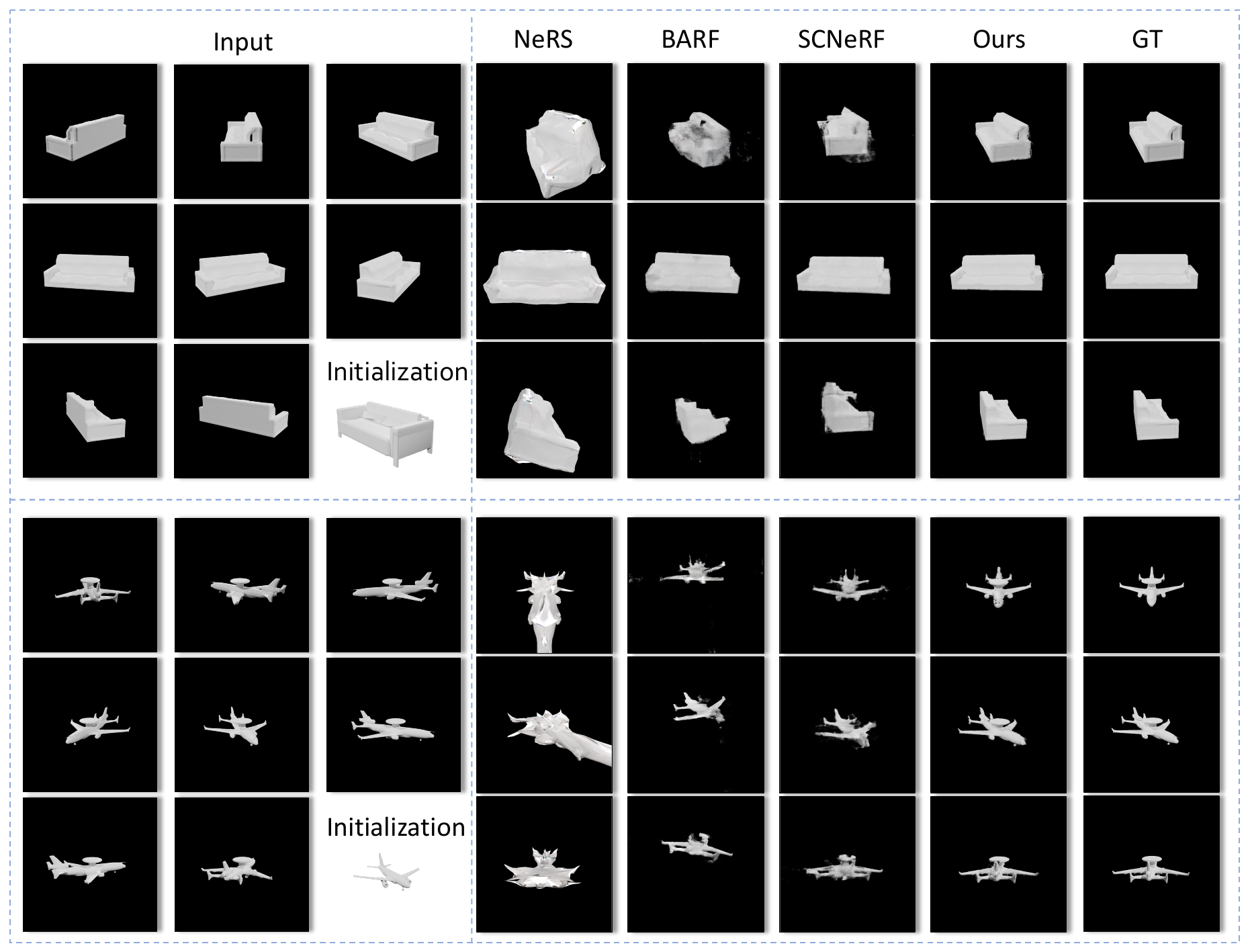}
  \caption{Qualitative comparison on ModelNet. On the left, it shows input images and the retrieved CAD model. On the right, there are novel views by NeRS, BARF, SCNeRF, ours and GT.}
  \label{newtu}
\end{figure*}

\begin{figure*}[h]
\centering
  \includegraphics[width=1.0\linewidth]{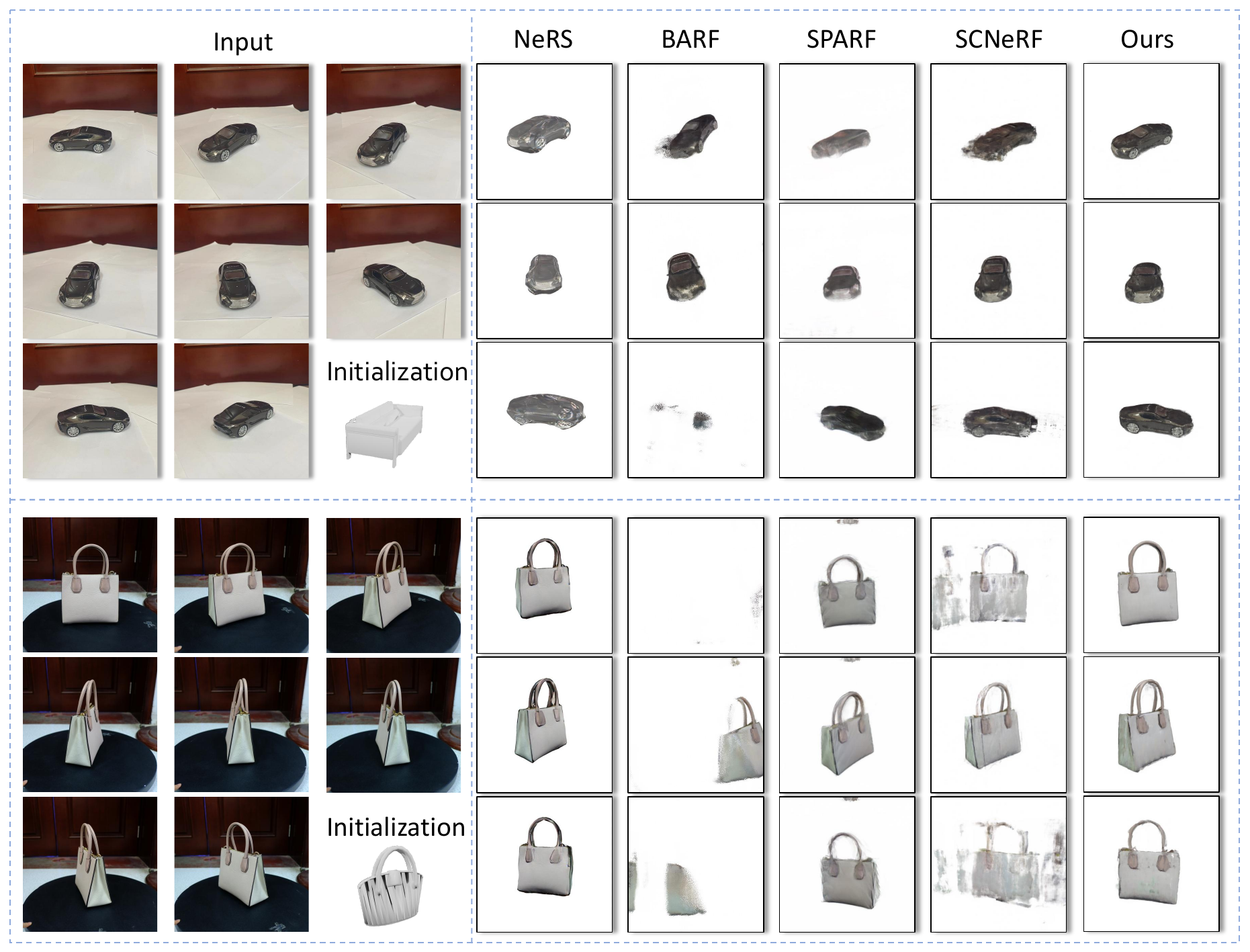}
  \caption{Qualitative comparison on real data. On the left, it shows input images and the retrieved CAD model. On the right, there are novel views by NeRS, BARF, SPARF, SCNeRF and ours.}
  \label{comp2}
\end{figure*}

\begin{figure*}[t]
\centering
  \includegraphics[width=0.9\linewidth]{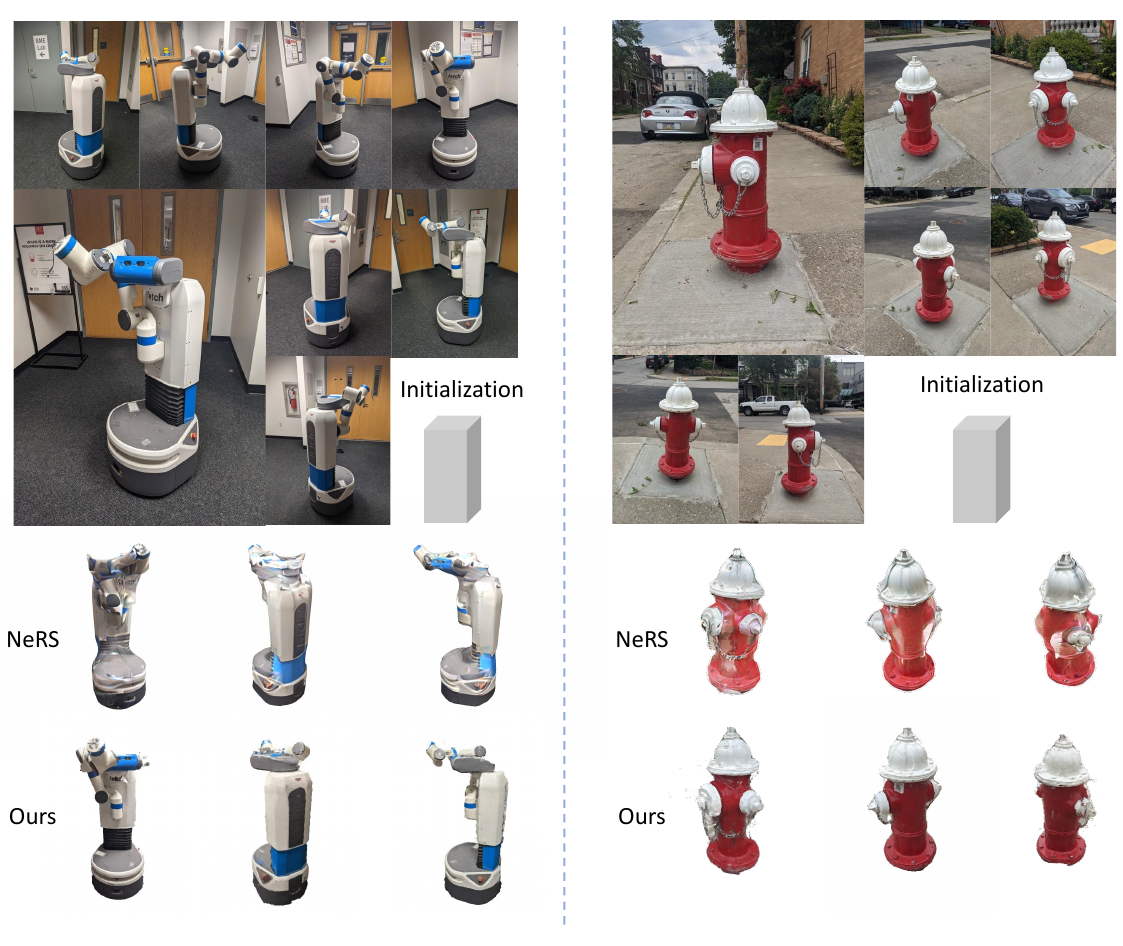} 
  \caption{Qualitative comparison of novel view synthesis to NeRS on two real objects from NeRS. Here we use the initial mesh and poses provided in their dataset for both methods. Input images and initial meshes are shown on the top. NeRS uses cuboids of various sizes as shape initialization for all data. The results of three novel views are shown at the bottom. With their initialization, our method could also get accurate reconstructions and more accurate shape details without their smooth constraints on surfaces.}
  \label{ners}
\end{figure*}

\begin{figure}[h]
\centering
  \includegraphics[width=1.0\linewidth]{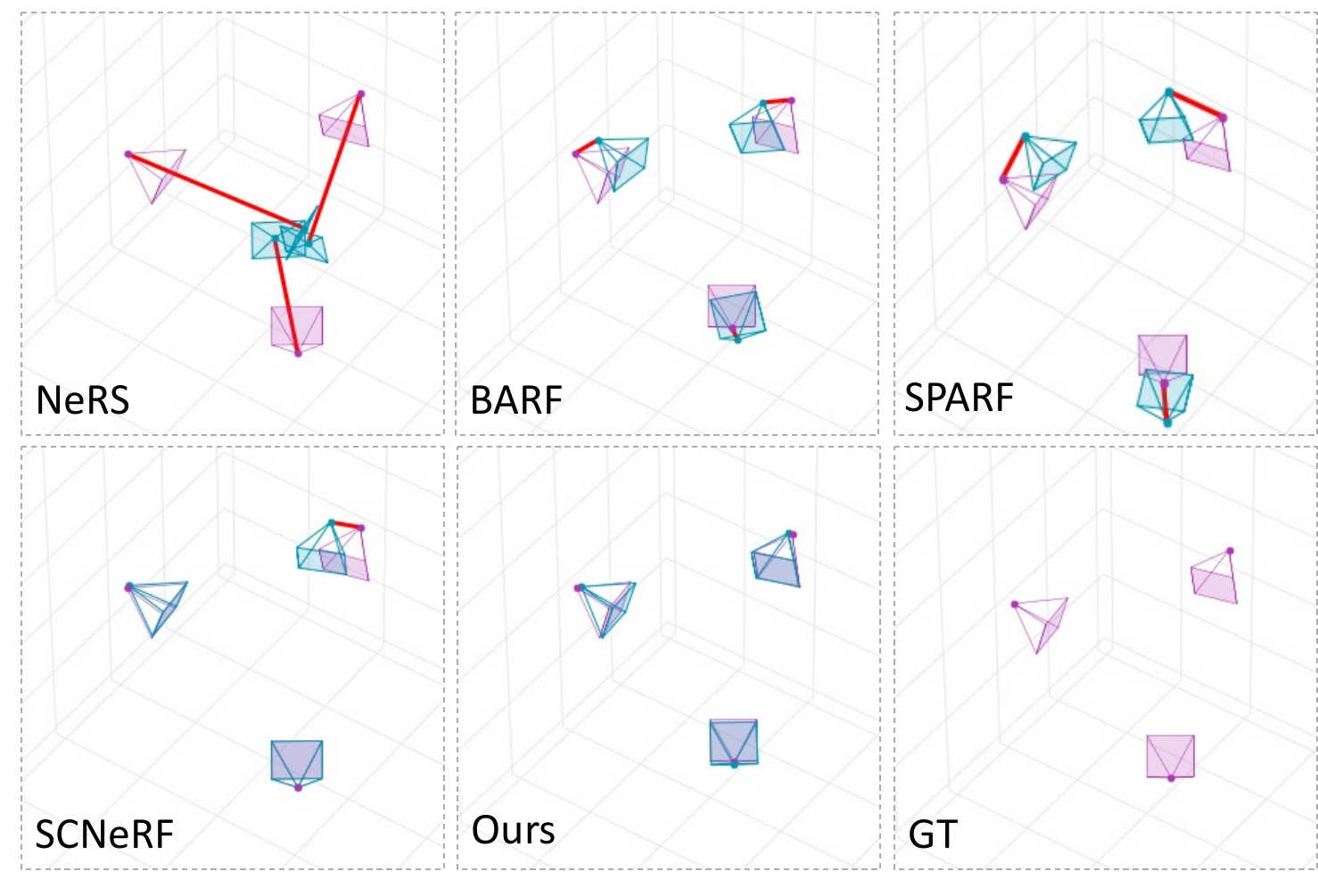}
  \caption{We show the registration results of the optimized poses (color in blue) with the ground-truth poses (color in pink) on the ``lamp" data for 3 input images.}
  \label{poses}
\end{figure}


\subsection{Training images with backgrounds}

CAD-NeRF is aimed at reconstructing objects from few-view images, while it still retains the ability to reconstruct scenes as the original NeRF, i.e. object images with backgrounds. 
The pipeline of training images with backgrounds is shown in Figure~\ref{tu}, with two phases. Specifically, the first phase (first 5k iterations) focuses on the object, with the retrieved poses and model, it learns the initial density field, the loss function $\mathcal{L}_{\text{density}}$ used is the same as in Equation \ref{ldensity}. In the second phase (after 5000 iterations), points from both the object and backgrounds are sampled, and fed into the deformation, the density and the color network, using the $\mathcal{L}_{\text {total }}$ loss in Equation \ref{lall} to jointly optimize the color and density until the network converges. As introduced in Section 3.4, the pose is optimized between 7.5k and 1w iterations, and after 1w iterations, the pose is fixed to prevent overfitting.


%% file: results.tex




\section{Experiments}
\label{sec:experiments}
\subsection{Dataset and implementation details}

\ptitle{Dataset.}
The main dataset used in this experiment is ShapeNet~\cite{chang2015shapenet}, ModelNet~\cite{Wu_2015_CVPR} and captured real-world data. We use Blender~\cite{denninger2020blenderproc} to render the obj files from ShapeNet and ModelNet. 
New objects that are not in the mini library are rendered for evaluation, where 3/6/9 images are randomly selected from each dataset as training data, and the remaining views are used for testing. For real-world data, we place the objects on a tabletop and randomly captured 8-9 images. There are no distance and angle requirements for capturing. 
By retrieving from our mini library, we get the initialized model and poses. We test on 5 real objects, including a car, a handbag, a cup, a toy airplane, and an earphone. The results of two objects are in the main paper, and the others are in \textit{Appendixes}. 
Note that, for these data, we do not record any intrinsic and extrinsic parameters of the camera, and they are all initialized by the intrinsic and extrinsic of matched images in the mini library. It also greatly increases the application scenarios of our approach. In addition, we also used the real dataset published by the NeRS~\cite{zhang2021ners} for more comparisons, which has less than 9 images for each object. 

\ptitle{Implementation details.}
The implementation uses the PyTorch architecture \cite{paszke2019pytorch}. Our network design is based on the NeRF\cite{2021nerf}, the color part remains 4 layer MLPs and density part adopts 4 ResNetBlockFC instead of 4 MLPs, deform part are all implemented as MLPs as 
 \cite{deng2021deformed}. The network is trained on an RTX 4090 by Adam~\cite{kingma2014adam} optimizer with an initial learning rate of 0.0005, and the total training time is 6 hours. For hyperparameters, $\lambda_a$ is set as 10, and $\lambda_b$ is set as 0.1. For object data, backgrounds are automatically removed by \textit{BackgroundRemover} for model and pose retrieval, we can still keep the backgrounds in NeRF reconstruction. 
 

\doublerulesep 0.1pt
\begin{table*}[t]
\begin{footnotesize}
\centering
\captionsetup{justification=centering}
\caption{
    Evaluation of ShapeNet data with the initial identity poses obtained by our retrieval poses. For 3/6/9 input views, we report the mean of all metrics for five different shapes of data, here the novel views for the test are the same for each object. }
\label{tab:369}
\small
\begin{tabular}{ccccccccccccc}
\hline
\multirow{2}{*}{} & \multicolumn{3}{c}{PSNR $\uparrow$} & \multicolumn{3}{c}{SSIM $\uparrow$} & \multicolumn{3}{c}{LPIPS $\downarrow$} & \multicolumn{3}{c}{Average $\downarrow$} \\ \cline{2-13} 
                  & 3-view & 6-view & 9-view & 3-view & 6-view & 9-view & 3-view  & 6-view & 9-view & 3-view  & 6-view  & 9-view  \\ \hline
BARF              & 18.860 & 18.968 & 20.161 & 0.881  & 0.881  & 0.901  & 0.311   & 0.309  & 0.236  & 0.115   & 0.115   & 0.092   \\
NeRS              & 18.687 & 18.561 & 19.070 & 0.885  & 0.884  & 0.895  & 0.210   & 0.193  & 0.181  & 0.104   & 0.102   & 0.095   \\
SCNeRF            & 17.371 & 18.648 & 19.259 & 0.852  & 0.882  & 0.887  & 0.293   & 0.190  & 0.175  & 0.128   & 0.097   & 0.089   \\
SPARF             & 18.459 & 20.781 & 20.759 & 0.880  & \textbf{0.933}  & \textbf{0.933}  & 0.220   & 0.094  & 0.092  & 0.104   & 0.059   & 0.059   \\ \hline
\textbf{Ours}              & \textbf{21.137} & \textbf{21.669} & \textbf{21.931} & \textbf{0.918}  & 0.929  & 0.930  & \textbf{0.105}   & \textbf{0.082}  & \textbf{0.081}  & \textbf{0.067}   & \textbf{0.058}   & \textbf{0.056}   \\ \hline
\end{tabular}
\end{footnotesize}
\end{table*}

\ptitle{Metrics.}
Following \cite{deng2022depth}, we evaluate the performance with 3 metrics: the Peak Signal-to-Noise Ratio (PSNR) \cite{wang2004image}, the Structure Similarity Index Measure (SSIM) \cite{wang2004image} and the Learned Perceptual Image Patch Similarity (LPIPS) \cite{zhang2018unreasonable} on the RGB of novel views. We also report the Average metric following \cite{Yang_2023_CVPR}, which is calculated by the geometric mean of MSE = $10 ^{-\text{PSNR}/10}$, $\sqrt{1-\text{SSIM}}$, and LPIPS. Besides, we report the average rotation and translation errors for pose registration as \cite{lin2021barf,hamdi2022sparf} after the pose optimization stage.

\doublerulesep 0.1pt
\begin{table}[h]
\begin{footnotesize}
\caption{Evaluation of synthetic data on ModelNet with the initial identity poses obtained by our retrieval poses.} \label{tab:modelnet}
\begin{tabular*}{\columnwidth}{p{0.2\columnwidth}p{0.2\columnwidth}p{0.2\columnwidth}p{0.2\columnwidth}}
\toprule
Method & PSNR↑   & SSIM↑ & LPIPS↓ \\
\midrule
NeRS   & 11.285  & 0.818 & 0.306 \\
SCNeRF & 15.833   & 0.875 & 0.221 \\
BARF   & 16.763   & 0.891 & 0.242 \\
\textbf{Ours}   & \textbf{24.026}   & \textbf{0.964}  & \textbf{0.046}  \\ 
\bottomrule
\end{tabular*}
\end{footnotesize}
\end{table}

\doublerulesep 0.1pt
\begin{table}[h]
\begin{footnotesize}
\caption{The comparison of rotation errors of optimized poses for 3/6/9 input images, the training starts from initial identity poses by our retrieved. Here the rotation errors are in degrees.} \label{tab:shape-pose1}
\begin{tabular*}{\columnwidth}{p{0.2\columnwidth}p{0.2\columnwidth}p{0.2\columnwidth}p{0.2\columnwidth}}
\toprule
Method       & 3-view & 6-view & 9-view \\ 
\midrule
BARF   & 21.486 & 19.750 & 16.681 \\
NeRS   & 60.534 & 12.091 & 9.327 \\
SCNeRF & 5.943  & 7.083  & 8.252  \\
SPARF  & 21.470 & 4.876  & 6.086  \\ 
\textbf{Ours}   & \textbf{5.273}  & \textbf{3.960}  & \textbf{3.914}  \\
\bottomrule
\end{tabular*}
\end{footnotesize}
\end{table}



\doublerulesep 0.1pt
\begin{table}[h]
\begin{footnotesize}
\caption{The comparison of translation errors of optimized poses for 3/6/9 input images, the training starts from initial identity poses by our retrieved. Here the translation errors are multiplied by 100.} \label{tab:shape-pose2}
\begin{tabular*}{\columnwidth}{p{0.2\columnwidth}p{0.2\columnwidth}p{0.2\columnwidth}p{0.2\columnwidth}}
\toprule
Method       & 3-view   & 6-view & 9-view \\
\midrule
BARF   & 60.003   & 43.144 & 34.925 \\
NeRS   & 194.467  & 36.103 & 21.564 \\
SCNeRF & 19.607   & 23.402 & 27.795 \\
SPARF  & 18.500   & 9.920  & 14.660 \\ 
\textbf{Ours}   & \textbf{14.960}   & \textbf{8.140}  & \textbf{7.900}  \\ 
\bottomrule
\end{tabular*}
\end{footnotesize}
\end{table}



\begin{figure}[t]
\centering
  \includegraphics[width=1.0\linewidth]{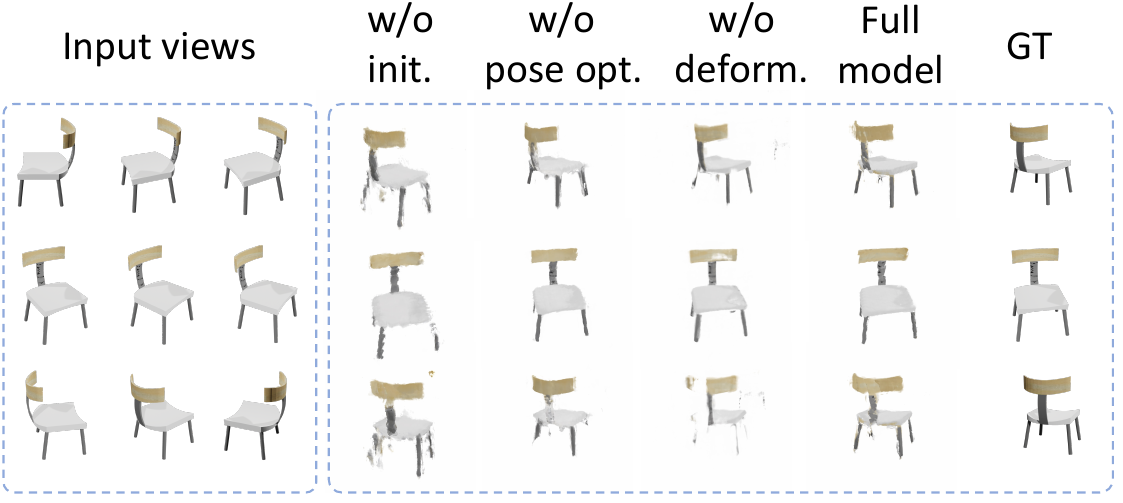}
  \caption{Ablations on Shapenet. Here ``w/o'' is short for ``without'', ``opt.'' is short for ``optimization'', and ``deform.'' means ``deformation ''. All input views are shown on the left. The right part shows the results of three ablations (without the initial mesh, without pose optimization, without deformation), our full model, and ground truth.}
  \label{aba}
\end{figure}

\begin{figure}[t]
\centering
  \includegraphics[width=1.0\linewidth]{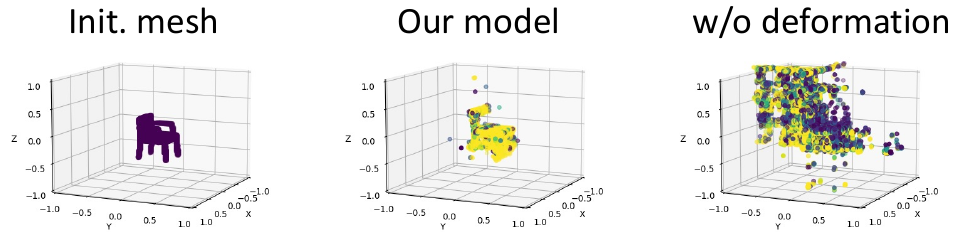}
  \caption{Ablation study of density field visualization. ``Init.'' is short for ``initial'', which means the initial mesh model from the retrieval module. Our learned density field is shown in the middle. ``w/o deformation'' is the result of the removed deformation module. }
  \label{abla density}
\end{figure}

\begin{figure}[t]
\centering
  \includegraphics[width=1.0\linewidth]{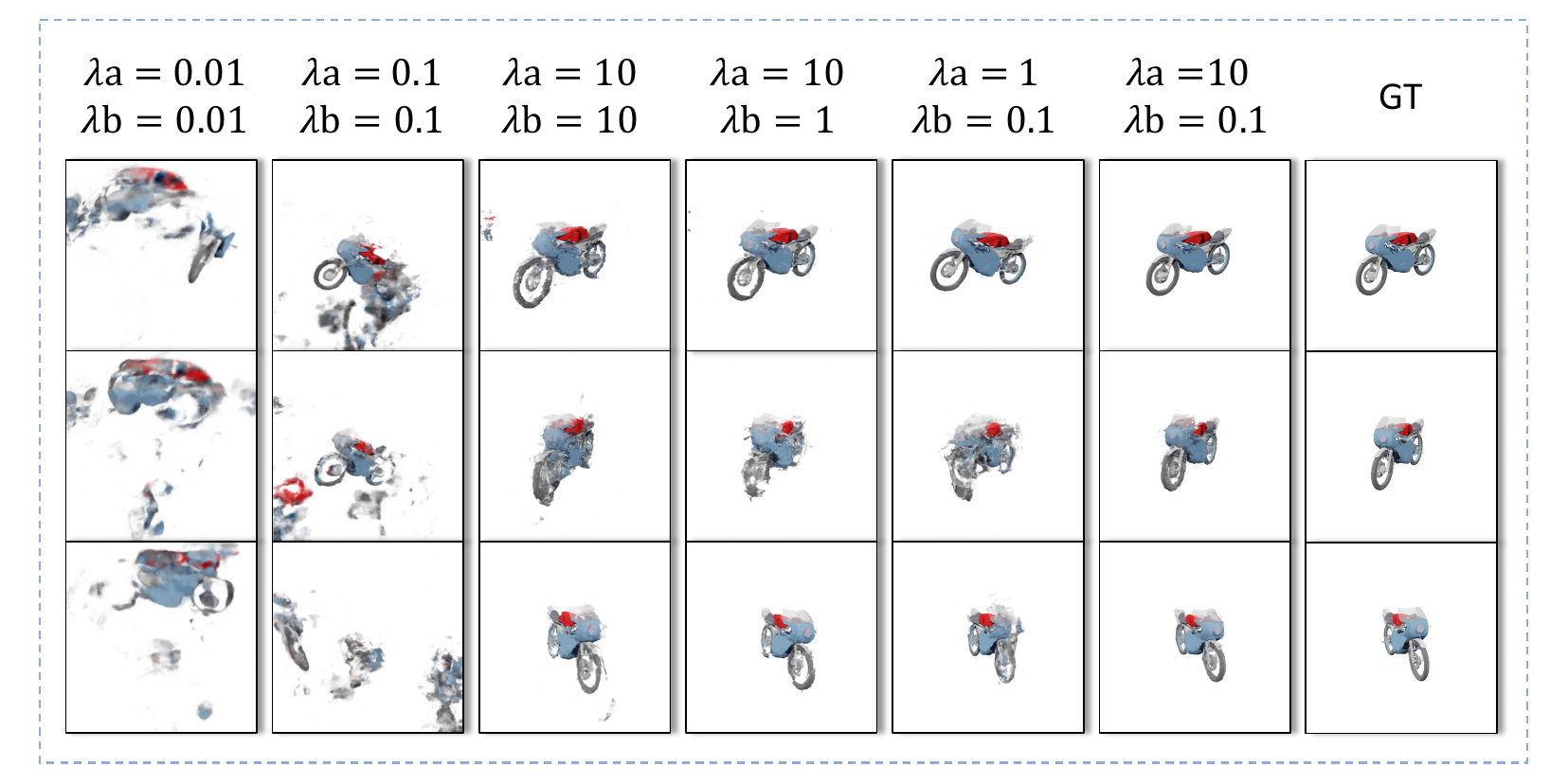}
  \caption{Ablation study of parameters setting.}
  \label{abla params}
\end{figure}

\subsection{Comparisons with state-of-the-arts}
\ptitle{Novel view synthesis.} As reported in Table \ref{tab:369}, while testing on 3/6/9 input images respectively, our method achieves comparable or best performance with other state-of-the-art in most metrics on the ShapeNet data. For a fair comparison, we give our retrieved poses as initial poses for the other four methods. In particular, we observe that our model achieves much higher PSNR than the other methods. SSIM of SPARF \cite{hamdi2022sparf} is slightly higher for 6/9 input views, their performance drops when input image numbers decrease to 3. CAD-Nerf performs robustly even for such extreme cases. 
Furthermore, we illustrate qualitative performance on ShapeNet data in Figure \ref{comp1}. BARF \cite{lin2021barf} and SCNeRF \cite{jeong2021self} are prone to produce more artifacts on account of the sparse images, which degrades the performance of the model. From Figure \ref{comp1}, we find that the shape of NeRS \cite{zhang2021ners} is better than BARF by constricting the geometry of objects to watertight surfaces. This constraint helps NeRS \cite{zhang2021ners} perform better at clustered objects, such as the motorcycle. However, it may fail on the scattered objects or objects with ``holes'', such as the lamp in Figure \ref{comp1}. SPARF \cite{hamdi2022sparf} reconstructs shapes well, but there are floating points leading to noises in rendered images. More visualizations are in \textit{Appendix B}.



We conduct further experiments on another dataset ModelNet to validate our approach. Unlike ShapeNet, this dataset only contains meshes without textures. To facilitate a fair comparison, we standardize the background to black. The results are shown in Figure \ref{newtu} and Table \ref{tab:modelnet}. It is worth mentioning that we attempted to run this experiment on SPARF~\cite{hamdi2022sparf}, but it failed, possibly due to the lack of textures leading to insufficient features, which hindered feature matching. As illustrated in Figure \ref{newtu}, although NeRS \cite{zhang2021ners}, BARF \cite{lin2021barf}, and SCNeRF \cite{jeong2021self} could achieve some results, their performance was poor, with noise around the objects and the tendency to lose detailed information. This indicates that the absence of textures also poses challenges for training their models. In contrast, our method achieves good results even without textures. As can be seen from Table \ref{tab:modelnet}, our three metrics significantly surpass theirs.

We also evaluate the real dataset, as shown in Figure \ref{comp2}, the left boxes show input images and our retrieval model, the right part is the novel view synthesis of SOTAs and our method. We observe that NeRS cannot model objects of non-genus-zero topologies 
because of its watertight constraints. BARF~\cite{lin2021barf} does not work well on sparse view images. SPARF~\cite{hamdi2022sparf} relies on a pre-trained dense correspondence network to extract matches between the training views, it does not work with texture-less objects such as ``car''. With only 8-9 images, benefiting from the deformation module, our method gets high-quality novel views. 
Moreover, we place objects in the scene for training with backgrounds, the results are in \textit{Appendix A}.

To make a better comparison, we conduct experiments on the real dataset proposed by NeRS \cite{zhang2021ners}, which has less than 9 images for each object. As shown in Figure \ref{ners}, in this experiment, we use cuboid initialization and marked initial pose, similar to the setting in NeRS. As we can see from Figure \ref{ners}, our method can learn better shapes (e.g. the robot head) and textures (the appearance of the red fire hydrant). The use of volume rendering also makes the rendered image closer to the real image.


\ptitle{Pose accuracy.}
We report more quantitative results on ShapeNet data about the comparison of errors of optimized poses for 3/6/9 input images in Table \ref{tab:shape-pose1} and Table \ref{tab:shape-pose2}, and show the registration results of the optimized poses with the ground-truth poses on the ``lamp'' data for 3 input images in Figure \ref{poses}. We find that NeRS \cite{zhang2021ners} targets sparse-view inputs, but it usually needs more than 6 images. Its performance drops significantly for 3 views. BARF \cite{lin2021barf} and SPARF \cite{hamdi2022sparf} also do not perform well with 3 view inputs. Our method consistently achieves the best overall performance in pose optimization. 

\doublerulesep 0.1pt
\begin{table}[h]
\begin{footnotesize}
\caption{Ablation study on ShapeNet datasets in Figure ~\ref{aba}.} \label{tab:ablation_curve}
\begin{tabular*}{\columnwidth}{p{0.3\columnwidth}p{0.2\columnwidth}p{0.2\columnwidth}p{0.2\columnwidth}}
\toprule
Method & PSNR↑   & SSIM↑ & LPIPS↓ \\
\midrule
    w/o Initialization & 19.082 & 0.909 & 0.202 \\
    w/o Pose optimization & 21.739  & 0.946 & 0.101 \\
    w/o Deformation & 21.469   & \textbf{0.948} & 0.120 \\
    Full Model & \textbf{22.678} & {0.943} & \textbf{0.099} \\
\bottomrule
\end{tabular*}
\end{footnotesize}
\end{table}

\begin{table}[h]
\begin{footnotesize}
\caption{Quantitative evaluation on hyperparameters, and qualitative comparisons can be found in Figure \ref{abla params}.}
\label{tab:hyperparameter}
\begin{tabular*}{\columnwidth}{p{0.3\columnwidth}p{0.2\columnwidth}p{0.2\columnwidth}p{0.2\columnwidth}p{0.2\columnwidth}}
\toprule
hyperparameter & PSNR↑         & SSIM↑         & LPIPS↓      \\
\midrule
$\lambda_a$=$\lambda_b$=0.1  & 15.742 & 0.784 & 0.372   \\
$\lambda_a$=$\lambda_b$=0.01 & 14.456 & 0.752 & 0.478  \\
$\lambda_a$=$\lambda_b$=10   & 18.673 & 0.882 & 0.166  \\
$\lambda_a$=10 / $\lambda_b$=1  & 19.730  & 0.900 & 0.124  \\
$\lambda_a$=1 / $\lambda_b$=0.1   & 19.497 & 0.895 & 0.132  \\
$\lambda_a$=10/$\lambda_b$=0.1(ours) & \textbf{20.911} & \textbf{0.925} & \textbf{0.068}  \\
\bottomrule
\end{tabular*}
\end{footnotesize}
\end{table}
\subsection{Ablation studies}



We perform ablation studies to verify the effectiveness of
each module in Table \ref{tab:ablation_curve} and Figure~\ref{aba}. Without initialization, due to the small number of images, the network fails to accurately predict the density. It leads to more artifacts and blurriness, thus getting the worst results. In this work, the pose of the input views is obtained by the retrieval module. 
We also test removing pose optimization and use the initial pose to train the network. 
PSNR drops to $19.082$, which is the minimum value. In Figure \ref{aba}, if the pose optimization is not carried out, the experimental results will be affected, resulting in missing parts (e.g. legs of the chair) in the image. The ablation of removing the deformation module gets a similar SSIM compared with the full model, although the PSNR is lower than the complete model. To further verify the effectiveness of the deformation module, we directly visualize the density field of the network in Figure \ref{abla density}. Here we show the original mesh retrieved in the first image on the left, the density field of our model is in the middle, and the density field without the deformation module is on the right. The deformation network optimizes the density through point coordinate translation and density correction, reduces the search space of density field learning, and guides the density from the initial shape to the target shape. The deformation module greatly contributes to density learning to reduce noise. 

Quantitative results of the different number of images on ShapeNet data for our method are shown in Table \ref{tab:369}-\ref{tab:shape-pose2}, $9$-view achieve the best PSNR $21.931$, and $9$-view get the comparable SSIM as $6$-view, measuring the similarity between two images. We find that as the number of images increases, the performance of the rendering views improves. However, the rotation errors and translation errors may not follow the same trend. This could be attributed to the fact that the additional images may introduce more errors, making the training more challenging. 

We conduct experiments on different selections of hyperparameter combinations $\lambda_a$ and $\lambda_b$ in Equation~\ref{lall}, and the results are shown in Figure \ref{abla params} and Table \ref{tab:hyperparameter}. We observe that changing the hyperparameter values $\lambda_a$ and $\lambda_b$ of the loss function produces different results, as reflected in both the visual quality of the images and the quantitative metrics. Through a series of experiments, we identify the optimal combination of $\lambda_a$ and $\lambda_b$ that balances reconstruction quality and perceptual quality. This combination is used in all our experiments and achieves the highest scores across three metrics PSNR, SSIM, LPIPS which are respectively $20.911$, $0.925$, and $0.068$. This combination can be explained by the fact that, for the initial density field, the displacement of 3D points is relatively large, thus the $\mathcal{L}_{\text{offset}}$ has a significant weight to regularize the offset. However, in the subsequent correction of the density field, the correction values are smaller, so the weight of $\mathcal{L}_{\text{correction }}$ is slightly less compared to the $\mathcal{L}_{color}$.



%% file: conclusion.tex
\section{Conclusions and Limitations}

In this paper, we present CAD-NeRF, a NeRF method for sparse-view uncalibrated images. We make use of priors from CAD datasets to retrieve initial shapes and poses, and jointly train density, deformation, poses, and textures. 
CAD-NeRF can be successfully applied to various shapes and performs well on both synthetic and real data. 
One limitation is that it requires order information of images, it may be a hard task if the input images are not few. We aim to get rid of this step in the future.  Another limitation is about the camera parameters, a hypothesis is that few-view images are taken from similar distances to the object as we render the CAD models in the library. We can solve it by sampling different camera positions, but it would take a longer time for retrieval. We will try to soften this constraint in the future. 

%% file: supp.tex
\appendix
\setcounter{table}{0}   
\setcounter{figure}{0}
\setcounter{section}{0}
\setcounter{equation}{0}
\renewcommand{\thetable}{A\arabic{table}}
\renewcommand{\thefigure}{A\arabic{figure}}
\renewcommand{\thesection}{Appendix \Alph{section}}
\renewcommand{\theequation}{A\arabic{equation}}

\section*{Appendixes}

We present additional experiment details of CAD-NeRF. 



\subsection*{Appendix A: Comparisons to the state-of-the-arts}

\begin{figure*}[t]
\centering
  \includegraphics[width=1.0\linewidth]{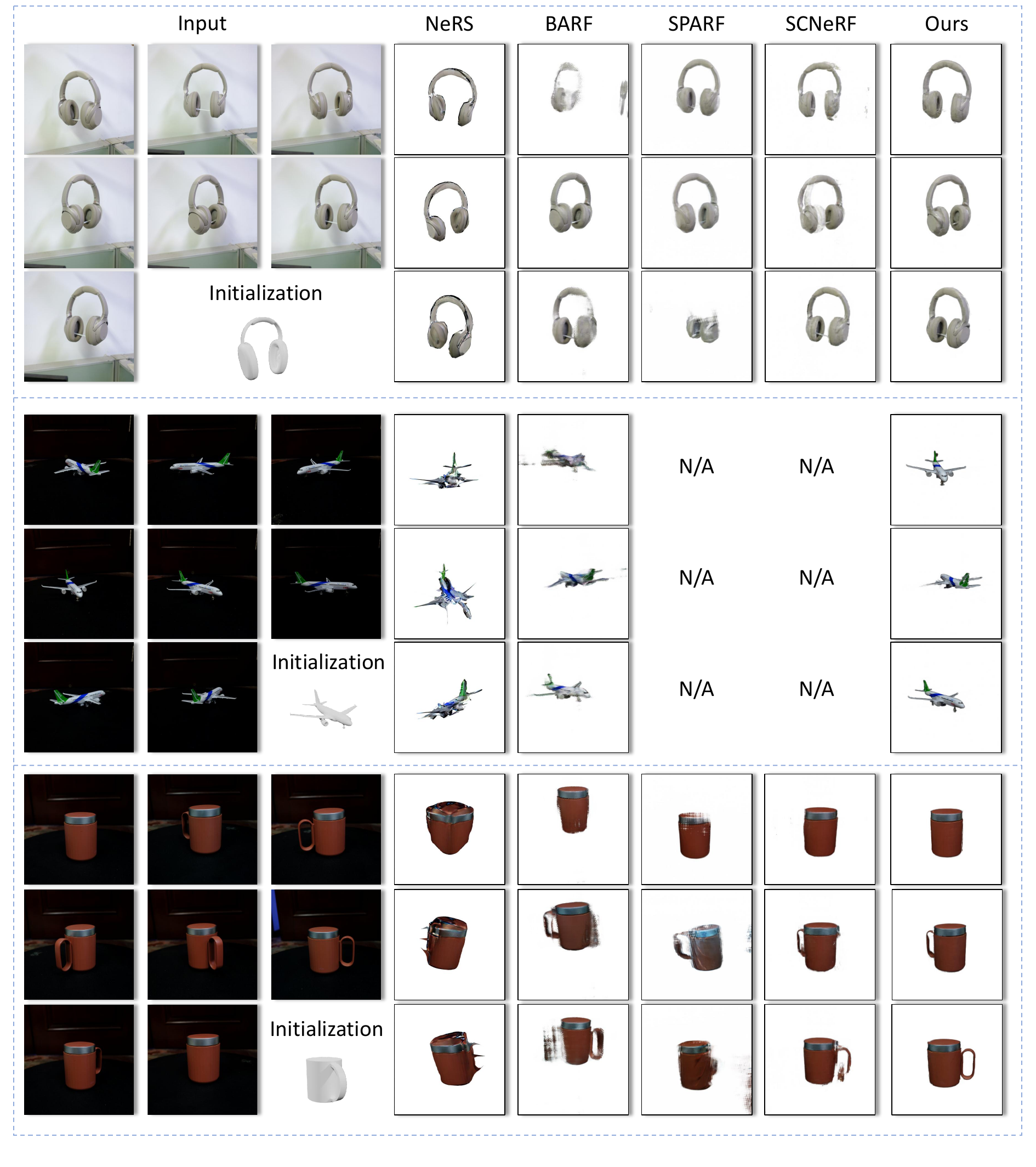}

  \caption{Comparison results on real data. The left shows the input images and the retrieved CAD model. On the right, it shows the results of novel view synthesis by NeRS, BARF, SPARF, SCNeRF and ours.}
  \label{real2}
\end{figure*}

We also capture a real dataset for evaluation. CAD-NeRF is compared with state-of-the-art methods on these real data. We demonstrate results on three objects, including a cup, an airplane toy, and a pair of earphones. The results are shown in Figure~\ref{real2}, the left boxes show the input images and the other columns are the novel view synthesis of different methods. 

As discussed in the main paper, NeRS~\cite{zhang2021ners} is not good at reconstructing scattered things because of its watertight constraints, so its results are prone to present black noises. BARF~\cite{lin2021barf} works well for normal cases of input numbers, but it is still challenging to work on sparse views. We find that SPARF~\cite{hamdi2022sparf} and SCNeRF~\cite{jeong2021self} work well on cases of rich textures/features, but fail in reconstructing the airplane because it has few features, which is not conducive to their matching network working. Different from these methods, CAD-NeRF performs better across these objects. 

\begin{figure*}[t]
\centering
  \includegraphics[width=0.9\linewidth]{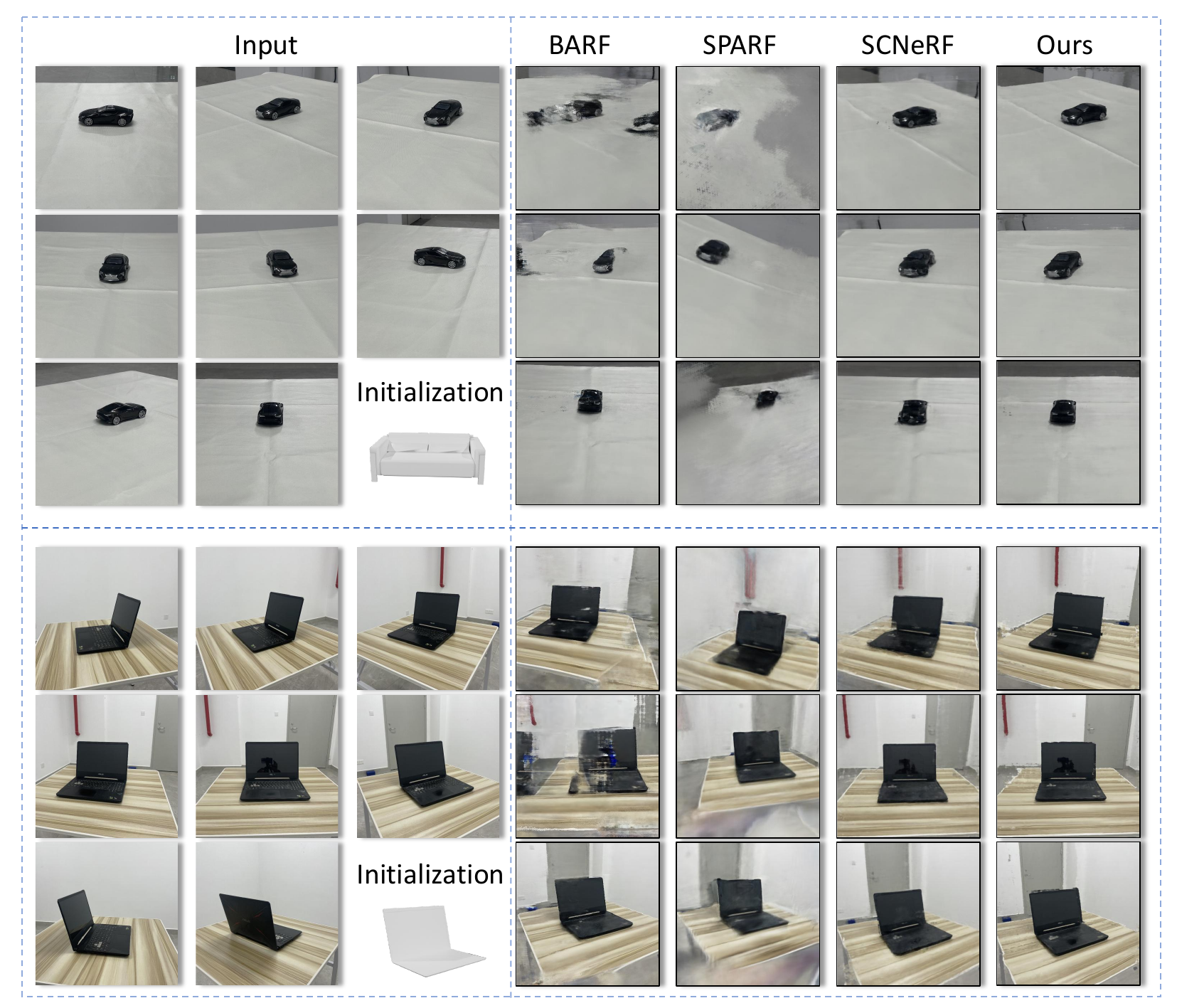}

  \caption{Comparison results on real scene data. The left shows the input images and the retrieved CAD model. On the right, it shows the results of novel view synthesis by BARF, SPARF, SCNeRF and ours.}
  \label{real1}

\end{figure*}

\begin{figure*}[t]
\centering
  \includegraphics[width=0.8\linewidth]{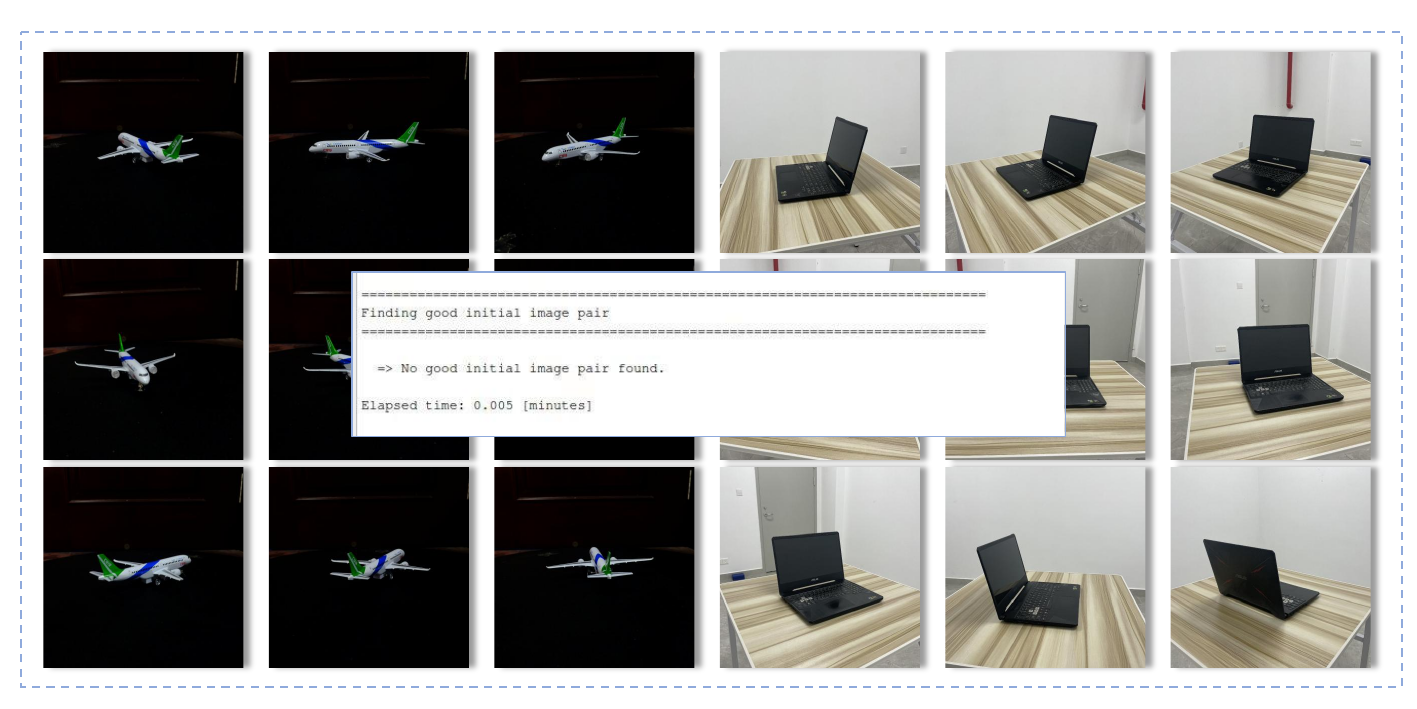}
  \caption{Examples of results that fail on COLMAP~\cite{schoenberger2016sfm} for training data.}
  \label{colmap}
\end{figure*}

Moreover, we place objects in the scene for training with backgrounds. Specifically, the objects are put on the table and take less than 9 images around the table, the background is fed into the network during training. We compare with several state-of-the-art methods on this real-scene dataset, as shown in Figure~\ref{real1}. All training starts from initial poses retrieved by our pipeline since COLMAP~\cite{schoenberger2016sfm} fails to work for sparse view cases. As shown in Figure~\ref{colmap},  the objects are smooth in textures and sparse in features, which makes COLMAP hard to initialize. BARF~\cite{lin2021barf} and SPARF~\cite{hamdi2022sparf} generate more noise because they require a more precise initial pose, and the pose we retrieved is a coarse one. SCNeRF~\cite{jeong2021self} achieves comparable results with CAD-NeRF, wherethey utilized Superglue model~\cite{sarlin2020superglue} for feature matching at the beginning of training, so it may fail in texture-less objects. From Figure~\ref{real1} we observe that dissimilar initial models also help and contribute to the results, such as ``sofa" to ``car". It is worth mentioning that we attempted to run Nope-NeRF~\cite{bian2023nope} on the scene dataset, 
as it aims to optimize the scene poses, but it failed to initialize since it needs more views and cannot work for such few-view scenarios.

Although CAD-NeRF is targeted at object images, but it works well for scene data, as long as we can find an object in the scene to serve as the probe for model and pose retrieval. The scene can be reconstructed with the poses retrieved by the target object. 

\subsection*{Appendix B: Ablation studies}

\begin{figure*}[t]
\centering
  \includegraphics[width=1.0\linewidth]{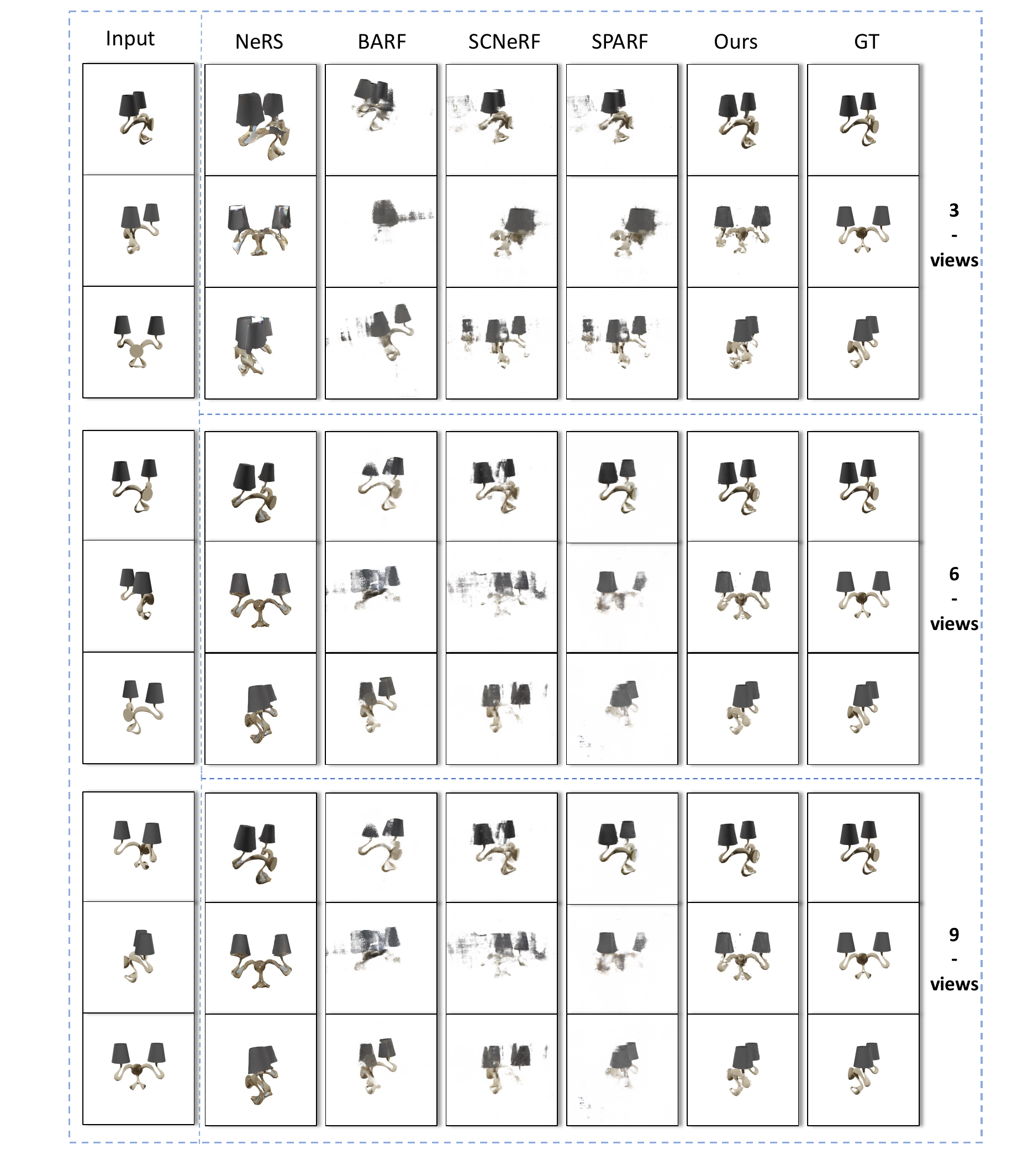}
  \caption{Ablation study about the number of input images for SOTAs and our proposed method on ``lamp" data. ``GT'' denotes ``ground truth''. All 9 input images are shown at the left part. ``3-views'' means only the first three rows of input images are used, and ``6-views'' means only the first six rows of input images are used. ``9-views'' means all images are used. It shows that even by reducing the number of input images to only three views, compared with SOTAs our method can still get reasonable results, showing the robustness of the method. }
  \label{368lamp}
\end{figure*}

\begin{figure*}[t]
\centering
  \includegraphics[width=1.0\linewidth]{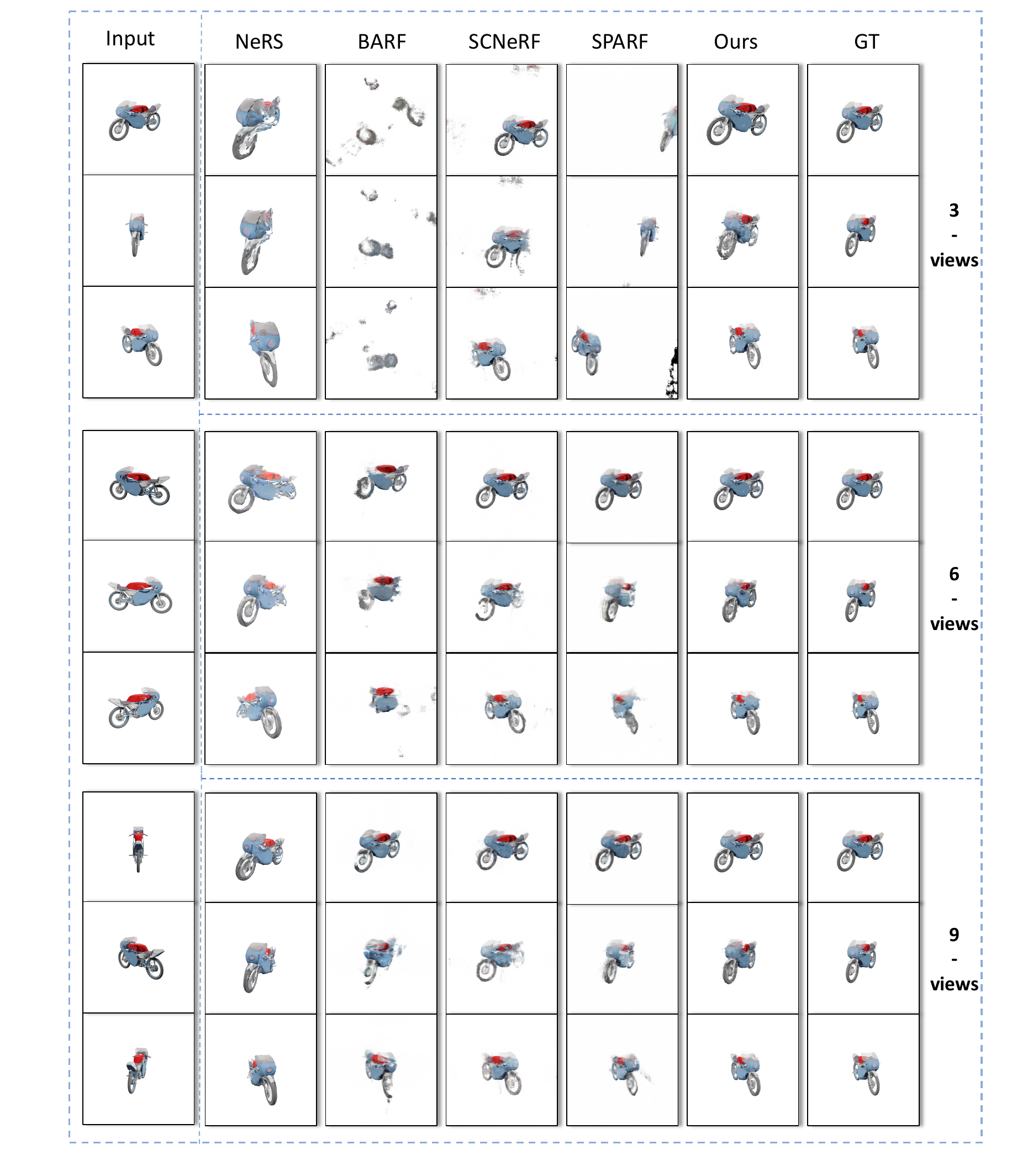}
  \caption{Ablation study about the number of input images for SOTAs and our proposed method on ``motorbike" data. ``GT'' denotes ``ground truth''. All 9 input images are shown at the left part. ``3-views'' means only the first three rows of input images are used, and ``6-views'' means only the first six rows of input images are used. ``9-views'' means all images are used. It shows that even by reducing the number of input images to only three views, compared with SOTAs our method can still get reasonable results, showing the robustness of the method. }
  \label{369moto}
\end{figure*}

\begin{figure*}[t]
\centering
  \includegraphics[width=1.0\linewidth]{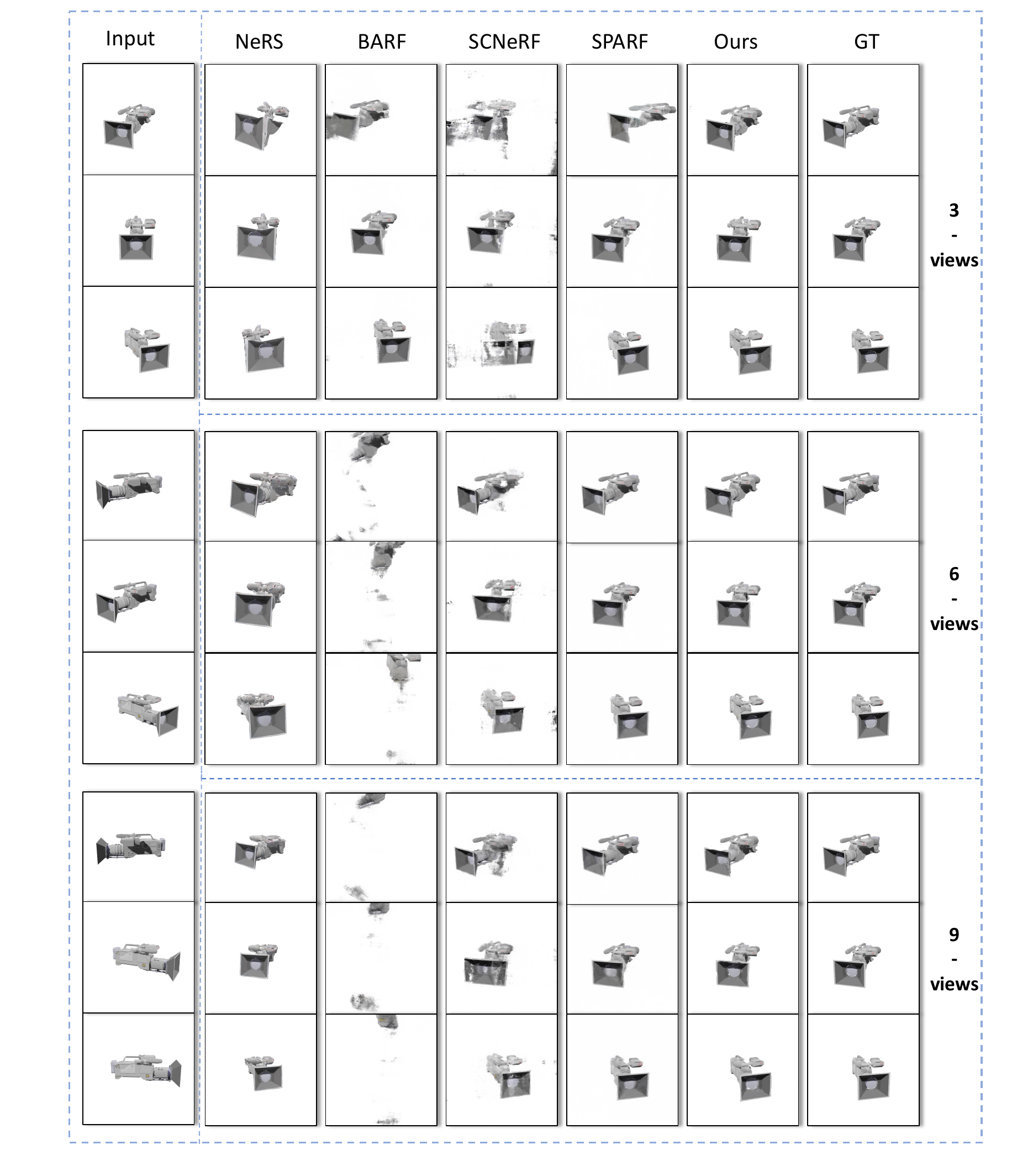}
  \caption{Ablation study about the number of input images for SOTAs and our proposed method on ``camera" data. ``GT'' denotes ``ground truth''. All 9 input images are shown at the left part. ``3-views'' means only the first three rows of input images are used, and ``6-views'' means only the first six rows of input images are used. ``9-views'' means all images are used. It shows that even by reducing the number of input images to only three views, compared with SOTAs our method can still get reasonable results, showing the robustness of the method. }
  \label{369cam}
\end{figure*}

\begin{figure*}[t]
\centering
  \includegraphics[width=1.0\linewidth]{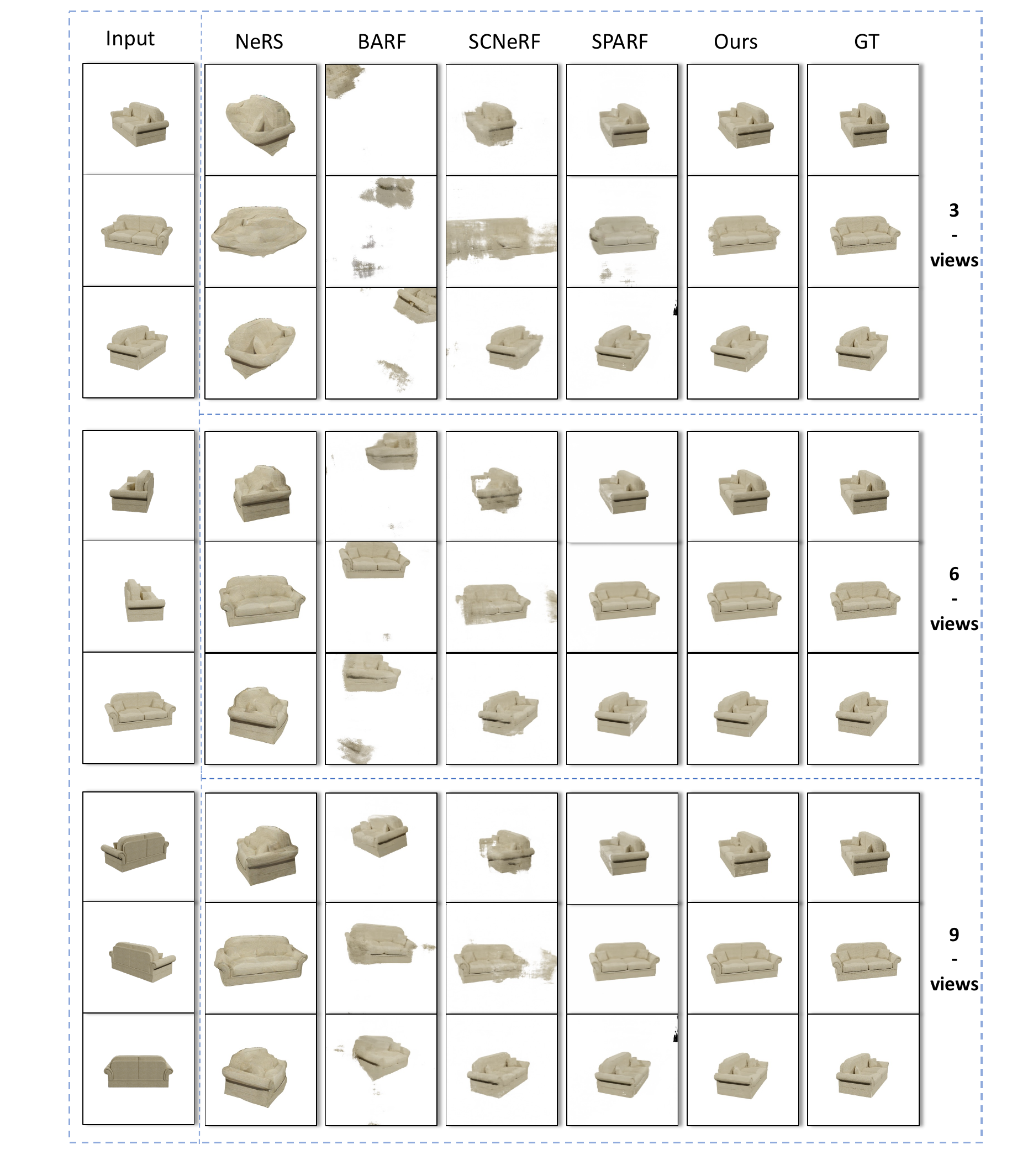}
  \caption{Ablation study about the number of input images for SOTAs and our proposed method on ``sofa" data. ``GT'' denotes ``ground truth''. All 9 input images are shown at the left part. ``3-views'' means only the first three rows of input images are used, and ``6-views'' means only the first six rows of input images are used. ``9-views'' means all images are used. It shows that even by reducing the number of input images to only three views, compared with SOTAs our method can still get reasonable results, showing the robustness of the method. }
  \label{369sofa}
\end{figure*}

\begin{figure*}[t]
\centering
  \includegraphics[width=1.0\linewidth]{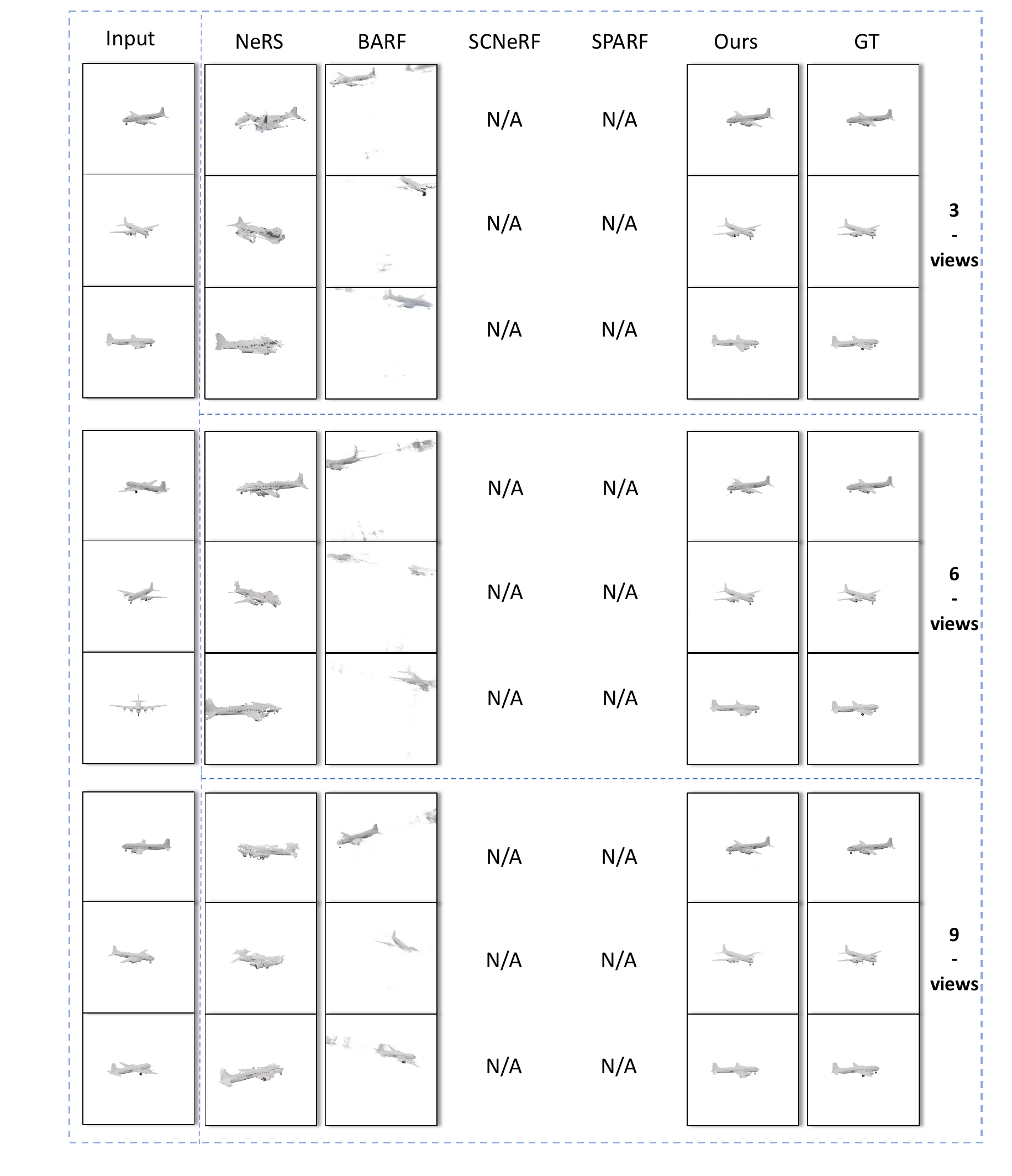}
  \caption{Ablation study about the number of input images for SOTAs and our proposed method on ``airplane" data. ``GT'' denotes ``ground truth''. All 9 input images are shown at the left part. ``3-views'' means only the first three rows of input images are used, and ``6-views'' means only the first six rows of input images are used. ``9-views'' means all images are used. It shows that even by reducing the number of input images to only three views, compared with SOTAs our method can still get reasonable results, showing the robustness of the method. }
  \label{369air}
\end{figure*}

Several ablation studies are presented in Figure~\ref{368lamp}-\ref{369air} about the number of input images, which are 3/6/9 views separately. To test the limit of methods, we reduce the number of images to $3$ views, $6$ views, compared with the original $9$ views. 

\begin{figure*}[t]
\centering
  \includegraphics[width=1.0\linewidth]{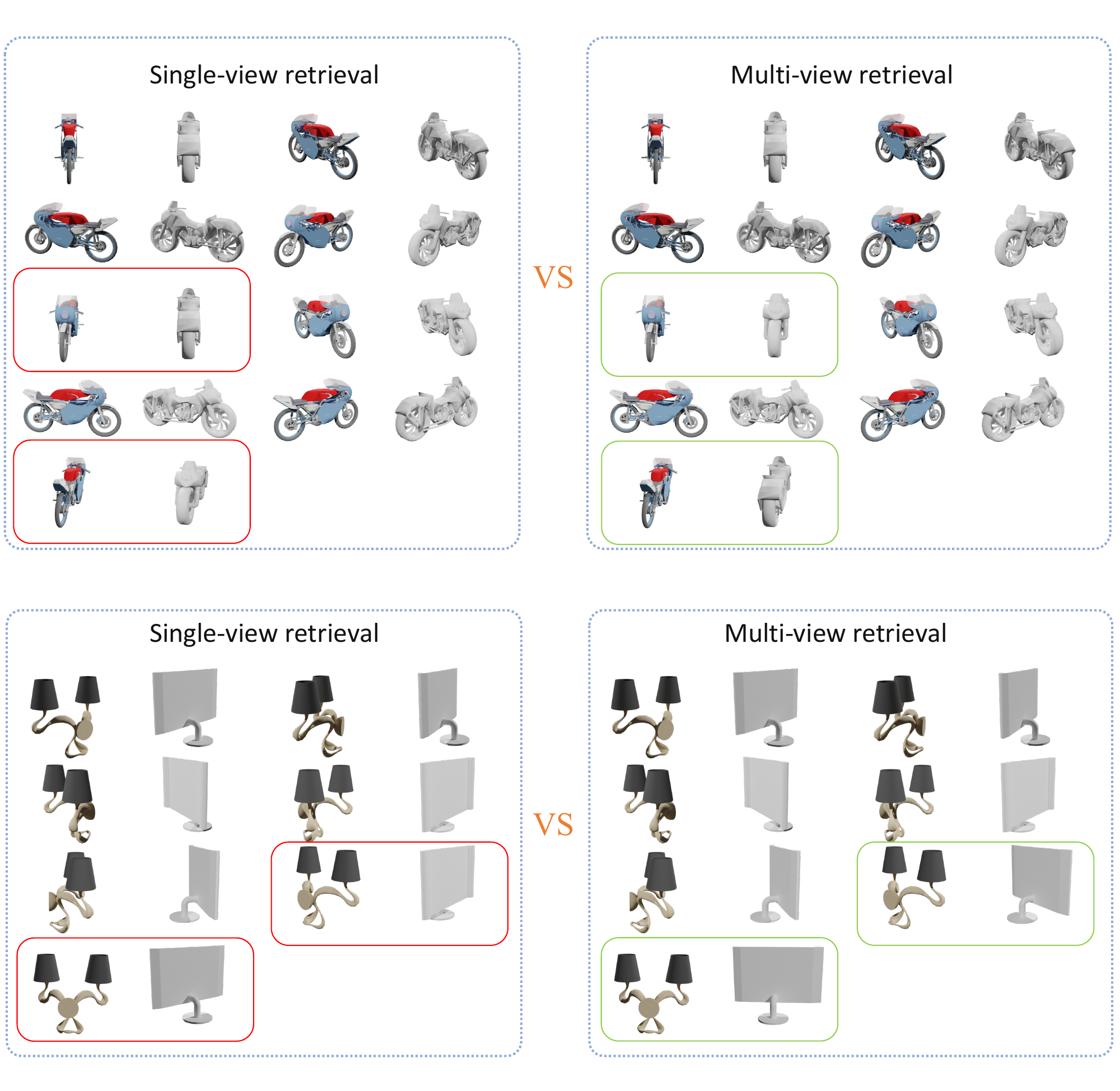}
  \caption{Comparisons of single-view retrieval and the proposed multi-view retrieval on ``motorbike" and ``lamp" data.}
  \label{lampcom}
\end{figure*}

\begin{figure*}[t]
\centering
  \includegraphics[width=1.0\linewidth]{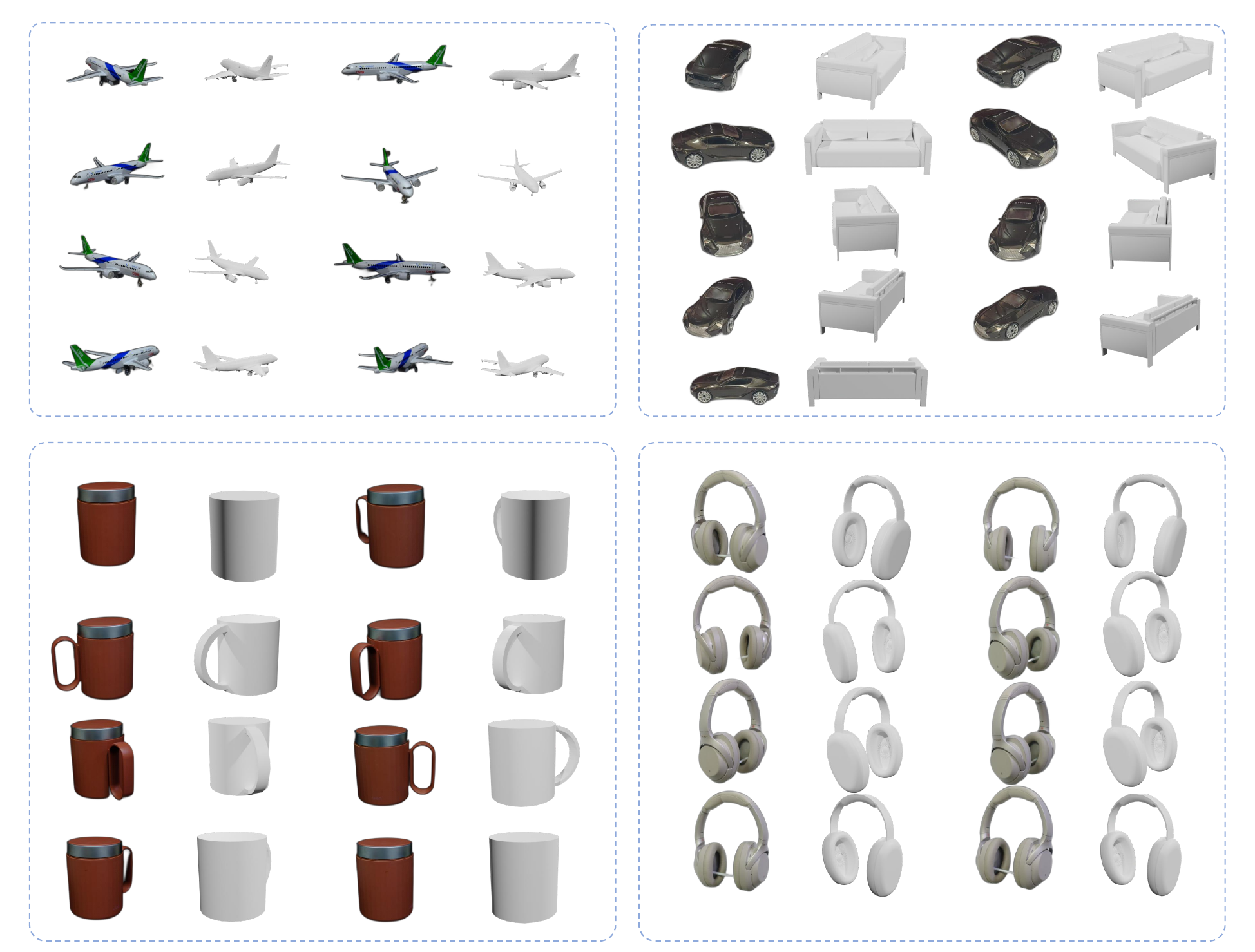}
  \caption{More multi-view retrieval results on real-world data.}
  \label{rs}
\end{figure*}

Even for only three views, i.e. only the first three rows of input images are used, CAD-NeRF still achieves a reasonable synthesis of the novel view. For $6$ views (the four to six rows of the input images), as the number of images increases, the noises gradually decrease and the novel views contain fewer artifacts. We also conduct this experiment on other methods, where BARF \cite{lin2021barf} and SPARF \cite{hamdi2022sparf} are more prone to noise when the number of images is less. Then SPARF~\cite{hamdi2022sparf} and SCNeRF \cite{jeong2021self} fail in the airplane data, which has too few image features. NeRS \cite{zhang2021ners} does not work for 3-views, which may lead to a very strange geometry, such as the ``sofa" data. 
In addition, we show more comparisons of single-view retrieval and the proposed multi-view retrieval. As shown in Figure \ref{lampcom}, the proposed multi-view retrieval may get more accurate camera poses. More multi-view retrieval results are presented in Figure \ref{rs}.

%% file: aaai25.bbl
\begin{thebibliography}{56}
\providecommand{\natexlab}[1]{#1}

\bibitem[{Barron et~al.(2021)Barron, Mildenhall, Tancik, Hedman, Martin-Brualla, and Srinivasan}]{barron2021mip}
Barron, J.~T.; Mildenhall, B.; Tancik, M.; Hedman, P.; Martin-Brualla, R.; and Srinivasan, P.~P. 2021.
\newblock Mip-nerf: A multiscale representation for anti-aliasing neural radiance fields.
\newblock In \emph{Proceedings of the IEEE/CVF International Conference on Computer Vision}, 5855--5864.

\bibitem[{Bian et~al.(2023)Bian, Wang, Li, Bian, and Prisacariu}]{bian2023nope}
Bian, W.; Wang, Z.; Li, K.; Bian, J.-W.; and Prisacariu, V.~A. 2023.
\newblock Nope-nerf: Optimising neural radiance field with no pose prior.
\newblock In \emph{Proceedings of the IEEE/CVF Conference on Computer Vision and Pattern Recognition}, 4160--4169.

\bibitem[{Chang et~al.(2015)Chang, Funkhouser, Guibas, Hanrahan, Huang, Li, Savarese, Savva, Song, Su et~al.}]{chang2015shapenet}
Chang, A.~X.; Funkhouser, T.; Guibas, L.; Hanrahan, P.; Huang, Q.; Li, Z.; Savarese, S.; Savva, M.; Song, S.; Su, H.; et~al. 2015.
\newblock Shapenet: An information-rich 3d model repository.
\newblock \emph{arXiv preprint arXiv:1512.03012}.

\bibitem[{Chen et~al.(2021)Chen, Xu, Zhao, Zhang, Xiang, Yu, and Su}]{chen2021mvsnerf}
Chen, A.; Xu, Z.; Zhao, F.; Zhang, X.; Xiang, F.; Yu, J.; and Su, H. 2021.
\newblock Mvsnerf: Fast generalizable radiance field reconstruction from multi-view stereo.
\newblock In \emph{Proceedings of the IEEE/CVF International Conference on Computer Vision}, 14124--14133.

\bibitem[{Cheng et~al.(2023)Cheng, Esteves, Jampani, Kar, Maji, and Makadia}]{cheng2023lu}
Cheng, Z.; Esteves, C.; Jampani, V.; Kar, A.; Maji, S.; and Makadia, A. 2023.
\newblock Lu-nerf: Scene and pose estimation by synchronizing local unposed nerfs.
\newblock In \emph{Proceedings of the IEEE/CVF International Conference on Computer Vision}, 18312--18321.

\bibitem[{Choy et~al.(2016)Choy, Xu, Gwak, Chen, and Savarese}]{choy20163d}
Choy, C.~B.; Xu, D.; Gwak, J.; Chen, K.; and Savarese, S. 2016.
\newblock 3d-r2n2: A unified approach for single and multi-view 3d object reconstruction.
\newblock In \emph{Computer Vision--ECCV 2016: 14th European Conference, Amsterdam, The Netherlands, October 11-14, 2016, Proceedings, Part VIII 14}, 628--644. Springer.

\bibitem[{Deng et~al.(2022)Deng, Liu, Zhu, and Ramanan}]{deng2022depth}
Deng, K.; Liu, A.; Zhu, J.-Y.; and Ramanan, D. 2022.
\newblock Depth-supervised nerf: Fewer views and faster training for free.
\newblock In \emph{Proceedings of the IEEE/CVF Conference on Computer Vision and Pattern Recognition}, 12882--12891.

\bibitem[{Deng, Yang, and Tong(2021)}]{deng2021deformed}
Deng, Y.; Yang, J.; and Tong, X. 2021.
\newblock Deformed implicit field: Modeling 3d shapes with learned dense correspondence.
\newblock In \emph{Proceedings of the IEEE/CVF Conference on Computer Vision and Pattern Recognition}, 10286--10296.

\bibitem[{Denninger et~al.(2020)Denninger, Sundermeyer, Winkelbauer, Olefir, Hodan, Zidan, Elbadrawy, Knauer, Katam, and Lodhi}]{denninger2020blenderproc}
Denninger, M.; Sundermeyer, M.; Winkelbauer, D.; Olefir, D.; Hodan, T.; Zidan, Y.; Elbadrawy, M.; Knauer, M.; Katam, H.; and Lodhi, A. 2020.
\newblock Blenderproc: Reducing the reality gap with photorealistic rendering.
\newblock In \emph{International Conference on Robotics: Sciene and Systems, RSS 2020}.

\bibitem[{Fan et~al.(2024)Fan, Dai, Seo, and He}]{fan2024revisit}
Fan, B.; Dai, Y.; Seo, Y.; and He, M. 2024.
\newblock A revisit of the normalized eight-point algorithm and a self-supervised deep solution.
\newblock \emph{Visual Intelligence}, 2(1): 3.

\bibitem[{Fan, Su, and Guibas(2017)}]{fan2017point}
Fan, H.; Su, H.; and Guibas, L.~J. 2017.
\newblock A point set generation network for 3d object reconstruction from a single image.
\newblock In \emph{Proceedings of the IEEE conference on computer vision and pattern recognition}, 605--613.

\bibitem[{Feng et~al.(2018)Feng, Wan, Xu, Chen, Li, and S{\'a}nchez}]{feng2018perceptual}
Feng, X.; Wan, W.; Xu, R. Y.~D.; Chen, H.; Li, P.; and S{\'a}nchez, J.~A. 2018.
\newblock A perceptual quality metric for 3D triangle meshes based on spatial pooling.
\newblock \emph{Frontiers of Computer Science}, 12: 798--812.

\bibitem[{Gao et~al.(2024)Gao, Yi, Zhu, Zhuang, Chen, and Xu}]{gao2024generic}
Gao, Z.; Yi, R.; Zhu, C.; Zhuang, K.; Chen, W.; and Xu, K. 2024.
\newblock Generic Objects as Pose Probes for Few-Shot View Synthesis.
\newblock \emph{arXiv preprint arXiv:2408.16690}.

\bibitem[{Girdhar et~al.(2016)Girdhar, Fouhey, Rodriguez, and Gupta}]{girdhar2016learning}
Girdhar, R.; Fouhey, D.~F.; Rodriguez, M.; and Gupta, A. 2016.
\newblock Learning a predictable and generative vector representation for objects.
\newblock In \emph{Computer Vision--ECCV 2016: 14th European Conference, Amsterdam, The Netherlands, October 11-14, 2016, Proceedings, Part VI 14}, 484--499. Springer.

\bibitem[{Gkioxari, Malik, and Johnson(2019)}]{gkioxari2019mesh}
Gkioxari, G.; Malik, J.; and Johnson, J. 2019.
\newblock Mesh r-cnn.
\newblock In \emph{Proceedings of the IEEE/CVF International Conference on Computer Vision}, 9785--9795.

\bibitem[{Hamdi, Ghanem, and Nie{\ss}ner(2022)}]{hamdi2022sparf}
Hamdi, A.; Ghanem, B.; and Nie{\ss}ner, M. 2022.
\newblock SPARF: Large-Scale Learning of 3D Sparse Radiance Fields from Few Input Images.
\newblock \emph{arXiv preprint arXiv:2212.09100}.

\bibitem[{Hou et~al.(2023)Hou, Luo, Qin, Shao, and Chen}]{hou2023fus}
Hou, J.; Luo, C.; Qin, F.; Shao, Y.; and Chen, X. 2023.
\newblock FuS-GCN: Efficient B-rep based graph convolutional networks for 3D-CAD model classification and retrieval.
\newblock \emph{Advanced Engineering Informatics}, 56: 102008.

\bibitem[{Jain, Tancik, and Abbeel(2021)}]{jain2021putting}
Jain, A.; Tancik, M.; and Abbeel, P. 2021.
\newblock Putting nerf on a diet: Semantically consistent few-shot view synthesis.
\newblock In \emph{Proceedings of the IEEE/CVF International Conference on Computer Vision}, 5885--5894.

\bibitem[{Jeong et~al.(2021)Jeong, Ahn, Choy, Anandkumar, Cho, and Park}]{jeong2021self}
Jeong, Y.; Ahn, S.; Choy, C.; Anandkumar, A.; Cho, M.; and Park, J. 2021.
\newblock Self-calibrating neural radiance fields.
\newblock In \emph{Proceedings of the IEEE/CVF International Conference on Computer Vision}, 5846--5854.

\bibitem[{Kim, Seo, and Han(2022)}]{kim2022infonerf}
Kim, M.; Seo, S.; and Han, B. 2022.
\newblock Infonerf: Ray entropy minimization for few-shot neural volume rendering.
\newblock In \emph{Proceedings of the IEEE/CVF Conference on Computer Vision and Pattern Recognition}, 12912--12921.

\bibitem[{Kingma and Ba(2014)}]{kingma2014adam}
Kingma, D.~P.; and Ba, J. 2014.
\newblock Adam: A method for stochastic optimization.
\newblock \emph{arXiv preprint arXiv:1412.6980}.

\bibitem[{Lin et~al.(2021)Lin, Ma, Torralba, and Lucey}]{lin2021barf}
Lin, C.-H.; Ma, W.-C.; Torralba, A.; and Lucey, S. 2021.
\newblock Barf: Bundle-adjusting neural radiance fields.
\newblock In \emph{Proceedings of the IEEE/CVF International Conference on Computer Vision}, 5741--5751.

\bibitem[{Lin et~al.(2022)Lin, Peng, Xu, Yan, Shuai, Bao, and Zhou}]{lin2022enerf}
Lin, H.; Peng, S.; Xu, Z.; Yan, Y.; Shuai, Q.; Bao, H.; and Zhou, X. 2022.
\newblock Efficient Neural Radiance Fields for Interactive Free-viewpoint Video.
\newblock In \emph{SIGGRAPH Asia Conference Proceedings}.

\bibitem[{Liu et~al.(2022)Liu, Wang, Zhang, Xu, Liu, Zhang, and Gao}]{liu2022efficient}
Liu, S.; Wang, S.; Zhang, P.; Xu, K.; Liu, X.; Zhang, C.; and Gao, F. 2022.
\newblock Efficient one-pass multi-view subspace clustering with consensus anchors.
\newblock In \emph{Proceedings of the AAAI Conference on Artificial Intelligence}, volume~36, 7576--7584.

\bibitem[{Martin-Brualla et~al.(2021)Martin-Brualla, Radwan, Sajjadi, Barron, Dosovitskiy, and Duckworth}]{nerfinwild}
Martin-Brualla, R.; Radwan, N.; Sajjadi, M.~S.; Barron, J.~T.; Dosovitskiy, A.; and Duckworth, D. 2021.
\newblock Nerf in the wild: Neural radiance fields for unconstrained photo collections.
\newblock In \emph{Proceedings of the IEEE/CVF Conference on Computer Vision and Pattern Recognition}, 7210--7219.

\bibitem[{Mildenhall et~al.(2021)Mildenhall, Srinivasan, Tancik, Barron, Ramamoorthi, and Ng}]{2021nerf}
Mildenhall, B.; Srinivasan, P.~P.; Tancik, M.; Barron, J.~T.; Ramamoorthi, R.; and Ng, R. 2021.
\newblock Nerf: Representing scenes as neural radiance fields for view synthesis.
\newblock \emph{Communications of the ACM}, 65(1): 99--106.

\bibitem[{Niemeyer et~al.(2022)Niemeyer, Barron, Mildenhall, Sajjadi, Geiger, and Radwan}]{niemeyer2022regnerf}
Niemeyer, M.; Barron, J.~T.; Mildenhall, B.; Sajjadi, M.~S.; Geiger, A.; and Radwan, N. 2022.
\newblock Regnerf: Regularizing neural radiance fields for view synthesis from sparse inputs.
\newblock In \emph{Proceedings of the IEEE/CVF Conference on Computer Vision and Pattern Recognition}, 5480--5490.

\bibitem[{Paszke et~al.(2019)Paszke, Gross, Massa, Lerer, Bradbury, Chanan, Killeen, Lin, Gimelshein, Antiga et~al.}]{paszke2019pytorch}
Paszke, A.; Gross, S.; Massa, F.; Lerer, A.; Bradbury, J.; Chanan, G.; Killeen, T.; Lin, Z.; Gimelshein, N.; Antiga, L.; et~al. 2019.
\newblock Pytorch: An imperative style, high-performance deep learning library.
\newblock \emph{Advances in neural information processing systems}, 32.

\bibitem[{Qin et~al.(2018)Qin, Gao, Peng, Wu, Shen, and Grudtsin}]{qin2018fine}
Qin, F.; Gao, N.; Peng, Y.; Wu, Z.; Shen, S.; and Grudtsin, A. 2018.
\newblock Fine-grained leukocyte classification with deep residual learning for microscopic images.
\newblock \emph{Computer methods and programs in biomedicine}, 162: 243--252.

\bibitem[{Qin et~al.(2022)Qin, Qiu, Gao, and Bai}]{qin20223d}
Qin, F.; Qiu, S.; Gao, S.; and Bai, J. 2022.
\newblock 3D CAD model retrieval based on sketch and unsupervised variational autoencoder.
\newblock \emph{Advanced Engineering Informatics}, 51: 101427.

\bibitem[{Reiser et~al.(2021)Reiser, Peng, Liao, and Geiger}]{reiser2021kilonerf}
Reiser, C.; Peng, S.; Liao, Y.; and Geiger, A. 2021.
\newblock KiloNeRF: Speeding up Neural Radiance Fields with Thousands of Tiny MLPs.
\newblock arXiv:2103.13744.

\bibitem[{Riegler et~al.(2017)Riegler, Ulusoy, Bischof, and Geiger}]{riegler2017octnetfusion}
Riegler, G.; Ulusoy, A.~O.; Bischof, H.; and Geiger, A. 2017.
\newblock Octnetfusion: Learning depth fusion from data.
\newblock In \emph{2017 International Conference on 3D Vision (3DV)}, 57--66. IEEE.

\bibitem[{Sarlin et~al.(2020)Sarlin, DeTone, Malisiewicz, and Rabinovich}]{sarlin2020superglue}
Sarlin, P.-E.; DeTone, D.; Malisiewicz, T.; and Rabinovich, A. 2020.
\newblock Superglue: Learning feature matching with graph neural networks.
\newblock In \emph{Proceedings of the IEEE/CVF conference on computer vision and pattern recognition}, 4938--4947.

\bibitem[{Sch\"{o}nberger and Frahm(2016)}]{schoenberger2016sfm}
Sch\"{o}nberger, J.~L.; and Frahm, J.-M. 2016.
\newblock Structure-from-Motion Revisited.
\newblock In \emph{Conference on Computer Vision and Pattern Recognition (CVPR)}.

\bibitem[{Sun, Wu, and Gao(2024)}]{sun2024recent}
Sun, J.-M.; Wu, T.; and Gao, L. 2024.
\newblock Recent advances in implicit representation-based 3d shape generation.
\newblock \emph{Visual Intelligence}, 2(1): 9.

\bibitem[{Sun, Zhang, and Miao(2024)}]{sun2024review}
Sun, Y.; Zhang, X.; and Miao, Y. 2024.
\newblock A review of point cloud segmentation for understanding 3D indoor scenes.
\newblock \emph{Visual Intelligence}, 2(1): 14.

\bibitem[{Tancik et~al.(2022)Tancik, Casser, Yan, Pradhan, Mildenhall, Srinivasan, Barron, and Kretzschmar}]{tancik2022block}
Tancik, M.; Casser, V.; Yan, X.; Pradhan, S.; Mildenhall, B.; Srinivasan, P.~P.; Barron, J.~T.; and Kretzschmar, H. 2022.
\newblock Block-nerf: Scalable large scene neural view synthesis.
\newblock In \emph{Proceedings of the IEEE/CVF Conference on Computer Vision and Pattern Recognition}, 8248--8258.

\bibitem[{Tancik et~al.(2021)Tancik, Mildenhall, Wang, Schmidt, Srinivasan, Barron, and Ng}]{tancik2021metanerf}
Tancik, M.; Mildenhall, B.; Wang, T.; Schmidt, D.; Srinivasan, P.~P.; Barron, J.~T.; and Ng, R. 2021.
\newblock Learned initializations for optimizing coordinate-based neural representations.
\newblock In \emph{Proceedings of the IEEE/CVF Conference on Computer Vision and Pattern Recognition}, 2846--2855.

\bibitem[{Tian et~al.(2023)Tian, Qin, Yi, Zhu, and Xu}]{tian2023tensorformer}
Tian, H.; Qin, Z.; Yi, R.; Zhu, C.; and Xu, K. 2023.
\newblock Tensorformer: Normalized Matrix Attention Transformer for High-quality Point Cloud Reconstruction.
\newblock \emph{arXiv e-prints}, arXiv--2306.

\bibitem[{Tian and Xu(2024)}]{tian2024surface}
Tian, H.; and Xu, K. 2024.
\newblock Surface Reconstruction from Point Clouds via Grid-based Intersection Prediction.
\newblock \emph{arXiv preprint arXiv:2403.14085}.

\bibitem[{Wang et~al.(2018)Wang, Zhang, Li, Fu, Liu, and Jiang}]{wang2018pixel2mesh}
Wang, N.; Zhang, Y.; Li, Z.; Fu, Y.; Liu, W.; and Jiang, Y.-G. 2018.
\newblock Pixel2mesh: Generating 3d mesh models from single rgb images.
\newblock In \emph{Proceedings of the European conference on computer vision (ECCV)}, 52--67.

\bibitem[{Wang et~al.(2021{\natexlab{a}})Wang, Wang, Genova, Srinivasan, Zhou, Barron, Martin-Brualla, Snavely, and Funkhouser}]{wang2021ibrnet}
Wang, Q.; Wang, Z.; Genova, K.; Srinivasan, P.~P.; Zhou, H.; Barron, J.~T.; Martin-Brualla, R.; Snavely, N.; and Funkhouser, T. 2021{\natexlab{a}}.
\newblock Ibrnet: Learning multi-view image-based rendering.
\newblock In \emph{Proceedings of the IEEE/CVF Conference on Computer Vision and Pattern Recognition}, 4690--4699.

\bibitem[{Wang et~al.(2004)Wang, Bovik, Sheikh, and Simoncelli}]{wang2004image}
Wang, Z.; Bovik, A.~C.; Sheikh, H.~R.; and Simoncelli, E.~P. 2004.
\newblock Image quality assessment: from error visibility to structural similarity.
\newblock \emph{IEEE transactions on image processing}, 13(4): 600--612.

\bibitem[{Wang et~al.(2021{\natexlab{b}})Wang, Wu, Xie, Chen, and Prisacariu}]{wang2021nerfmm}
Wang, Z.; Wu, S.; Xie, W.; Chen, M.; and Prisacariu, V.~A. 2021{\natexlab{b}}.
\newblock Ne{RF}$--$: Neural Radiance Fields Without Known Camera Parameters.
\newblock \emph{arXiv preprint arXiv:2102.07064}.

\bibitem[{Wu et~al.(2019)Wu, Li, Huang, and Liu}]{wu2019joint}
Wu, G.; Li, Y.; Huang, Y.; and Liu, Y. 2019.
\newblock Joint view synthesis and disparity refinement for stereo matching.
\newblock \emph{Frontiers of Computer Science}, 13: 1337--1352.

\bibitem[{Wu et~al.(2015)Wu, Song, Khosla, Yu, Zhang, Tang, and Xiao}]{Wu_2015_CVPR}
Wu, Z.; Song, S.; Khosla, A.; Yu, F.; Zhang, L.; Tang, X.; and Xiao, J. 2015.
\newblock 3D ShapeNets: A Deep Representation for Volumetric Shapes.
\newblock In \emph{Proceedings of the IEEE Conference on Computer Vision and Pattern Recognition (CVPR)}.

\bibitem[{Xu et~al.(2022)Xu, Xu, Philip, Bi, Shu, Sunkavalli, and Neumann}]{xu2022point}
Xu, Q.; Xu, Z.; Philip, J.; Bi, S.; Shu, Z.; Sunkavalli, K.; and Neumann, U. 2022.
\newblock Point-NeRF: Point-based Neural Radiance Fields.
\newblock \emph{arXiv preprint arXiv:2201.08845}.

\bibitem[{Yan et~al.(2024)Yan, Zhou, Wang, Gao, and Yang}]{yan2024dialoguenerf}
Yan, Y.; Zhou, Z.; Wang, Z.; Gao, J.; and Yang, X. 2024.
\newblock Dialoguenerf: Towards realistic avatar face-to-face conversation video generation.
\newblock \emph{Visual Intelligence}, 2(1): 24.

\bibitem[{Yang et~al.(2019)Yang, Huang, Hao, Liu, Belongie, and Hariharan}]{yang2019pointflow}
Yang, G.; Huang, X.; Hao, Z.; Liu, M.-Y.; Belongie, S.; and Hariharan, B. 2019.
\newblock Pointflow: 3d point cloud generation with continuous normalizing flows.
\newblock In \emph{Proceedings of the IEEE/CVF international conference on computer vision}, 4541--4550.

\bibitem[{Yang, Pavone, and Wang(2023)}]{Yang_2023_CVPR}
Yang, J.; Pavone, M.; and Wang, Y. 2023.
\newblock FreeNeRF: Improving Few-Shot Neural Rendering With Free Frequency Regularization.
\newblock In \emph{Proceedings of the IEEE/CVF Conference on Computer Vision and Pattern Recognition (CVPR)}, 8254--8263.

\bibitem[{Yu et~al.(2021{\natexlab{a}})Yu, Ye, Tancik, and Kanazawa}]{yu2021pixelnerf}
Yu, A.; Ye, V.; Tancik, M.; and Kanazawa, A. 2021{\natexlab{a}}.
\newblock pixelnerf: Neural radiance fields from one or few images.
\newblock In \emph{Proceedings of the IEEE/CVF Conference on Computer Vision and Pattern Recognition}, 4578--4587.

\bibitem[{Yu et~al.(2021{\natexlab{b}})Yu, Zheng, Huang, Guo, Sun, and Yu}]{yu2021framework}
Yu, Z.; Zheng, X.; Huang, F.; Guo, W.; Sun, L.; and Yu, Z. 2021{\natexlab{b}}.
\newblock A framework based on sparse representation model for time series prediction in smart city.
\newblock \emph{Frontiers of Computer Science}, 15: 1--13.

\bibitem[{Yuan et~al.(2022)Yuan, Lai, Huang, Kobbelt, and Gao}]{yuan2022neural}
Yuan, Y.-J.; Lai, Y.-K.; Huang, Y.-H.; Kobbelt, L.; and Gao, L. 2022.
\newblock Neural radiance fields from sparse rgb-d images for high-quality view synthesis.
\newblock \emph{IEEE Transactions on Pattern Analysis and Machine Intelligence}, 45(7): 8713--8728.

\bibitem[{Zhang et~al.(2021)Zhang, Yang, Tulsiani, and Ramanan}]{zhang2021ners}
Zhang, J.; Yang, G.; Tulsiani, S.; and Ramanan, D. 2021.
\newblock NeRS: neural reflectance surfaces for sparse-view 3D reconstruction in the wild.
\newblock \emph{Advances in Neural Information Processing Systems}, 34: 29835--29847.

\bibitem[{Zhang et~al.(2018)Zhang, Isola, Efros, Shechtman, and Wang}]{zhang2018unreasonable}
Zhang, R.; Isola, P.; Efros, A.~A.; Shechtman, E.; and Wang, O. 2018.
\newblock The unreasonable effectiveness of deep features as a perceptual metric.
\newblock In \emph{Proceedings of the IEEE conference on computer vision and pattern recognition}, 586--595.

\bibitem[{Zhu et~al.(2024)Zhu, Yi, Wen, Zhu, and Xu}]{zhu2024relighting}
Zhu, X.; Yi, R.; Wen, X.; Zhu, C.; and Xu, K. 2024.
\newblock Relighting Scenes with Object Insertions in Neural Radiance Fields.
\newblock \emph{arXiv preprint arXiv:2406.14806}.

\end{thebibliography}
